\documentclass[sigconf, nonacm]{acmart}
\newcommand\vldbdoi{XX.XX/XXX.XX}

\newcommand\vldbavailabilityurl{https://github.com/fasttrack-nn/panorama}

\usepackage{algorithm}
\usepackage{tabularx}
\usepackage{algorithm}
\usepackage[noend]{algpseudocode}
\usepackage{amsthm}
\usepackage{booktabs}
\usepackage{enumitem}
\usepackage{graphicx}
\usepackage{float}
\usepackage{cleveref}
\usepackage{hyphenat}
\usepackage{subcaption}
\usepackage{wrapfig}
\usepackage{xfrac}
\usepackage{xspace}
\usepackage{enumitem}
\usepackage{listings}

\usepackage{microtype}

\definecolor{snippetKey}{rgb}{0,0,0}
\definecolor{snippetType}{rgb}{0,0,0}
\definecolor{snippetIntr}{rgb}{0,0,0}
\definecolor{snippetCom}{rgb}{0,0,0}
\lstdefinestyle{intrinsic}{%
  basicstyle=\ttfamily\footnotesize,
  language=C,
  classoffset=0,
  keywordstyle=\color{snippetKey}\bfseries,
  classoffset=1,
  morekeywords={__m128i,__m256i,__m256,__m512,__m512i,uint8_t,uint16_t,uint32_t,uint64_t,size_t},
  keywordstyle=\color{snippetType},
  classoffset=2,
  morekeywords={_pext_u64,_mm_loadu_si128,_mm_loadu_si256,_mm256_cvtepu8_epi32,_mm256_permutevar8x32_epi32,_mm512_loadu_ps,_mm512_add_ps,_mm512_cvtepu8_epi32,_mm512_i32gather_ps},
  keywordstyle=\color{snippetIntr}\bfseries,
  classoffset=0,
  commentstyle=\itshape\color{snippetCom},
  showstringspaces=false,
  frame=single,
  rulecolor=\color{snippetFrame},
  framerule=0.4pt,
  framesep=4pt,
  xleftmargin=0.5em,
  xrightmargin=0.5em,
  aboveskip=5pt,
  belowskip=5pt,
  breaklines=true,
  breakatwhitespace=true,
  postbreak=\mbox{\textcolor{snippetFrame}{$\hookrightarrow$}\space},
}
\setlist[itemize]{leftmargin=*}
\setlist[enumerate]{leftmargin=*}

\usepackage[textsize=small,backgroundcolor=yellow!20,linecolor=orange,bordercolor=orange]{todonotes}

\newcommand{\pkv}[1]{#1} 
\newcommand{\pkvall}[1]{#1} 
\newcommand{\pk}[1]{#1}

\newcommand{\rev}[1]{#1}
\newcommand{\akash}[1]{#1}
\newcommand{\alexis}[1]{#1}
\newcommand{\vansh}[1]{#1}

\makeatletter
\renewcommand*{\@fnsymbol}[1]{\ensuremath{\ifcase#1\or
  \dagger\or \ddagger\or \mathsection\or \mathparagraph\or
  \|\or **\or \dagger\dagger\or \ddagger\ddagger\else\@ctrerr\fi}}
\makeatother

\newtheorem{problem}{Problem}

\newcommand{\anns}{\textsc{ANNS}\xspace}
\newcommand{\name}{\textsc{Panorama}\xspace}

\newcommand{\faiss}{\textsc{Faiss}\xspace}
\newcommand{\ivfflat}{\textsc{IVFFlat}\xspace}

\newcommand{\ivfpq}{\textsc{IVFPQ}\xspace}

\newcommand{\hnsw}{\textsc{HNSW}\xspace}

\newcommand{\bigO}{\ensuremath{\mathcal{O}}}
\newcommand{\ub}{\ensuremath{\mathsf{UB}}\xspace}
\newcommand{\lb}{\ensuremath{\mathsf{LB}}\xspace}

\newcommand{\kub}{\ensuremath{d_k}\xspace}
\newcommand{\kth}{\ensuremath{k^\mathrm{th}}\xspace}
\newcommand{\query}{\ensuremath{\mathbf{q}}\xspace}

\newcommand{\candidates}{\ensuremath{\Gamma}\xspace}
\newcommand{\dataset}{\ensuremath{\mathcal{D}}\xspace}

\newcommand{\xvec}{\ensuremath{\mathbf{x}}\xspace}

\newcommand{\svec}{\ensuremath{\mathbf{s}}\xspace}

\newcommand{\sumell}{\ensuremath{\sum_{j=m_{\ell_1}+1}^{m_{\ell_2}}}\xspace}

\newcommand{\margin}{\ensuremath{\Delta}\xspace}
\newcommand{\speedup}{$28.9\times$}

\newcommand{\panoramaTableStyle}{%
  \small%
  \setlength{\tabcolsep}{5pt}%
  \renewcommand{\arraystretch}{1.0}%
}

\newcommand{\figLevelBatch}{%
  \begin{figure}[t]
    \centering
    \includegraphics[width=\columnwidth]{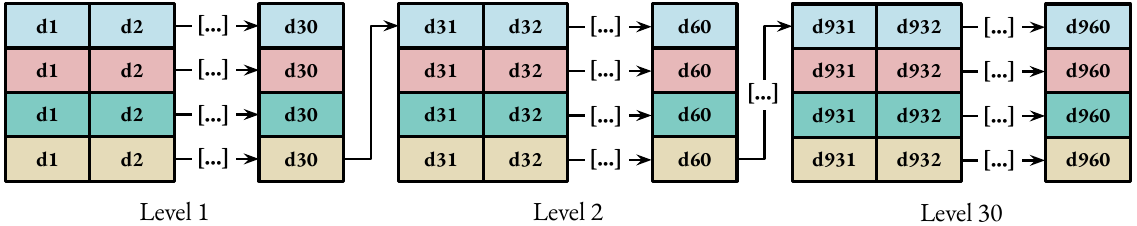}
    \caption{Batch 1 storage for \ivfflat~on GIST.}
    \label{fig:levelbatch}
  \end{figure}%
}

\newcommand{\figStorage}{%
  \begin{figure}[t]
    \vspace{-2mm}
    \centering
    \includegraphics[width=\columnwidth]{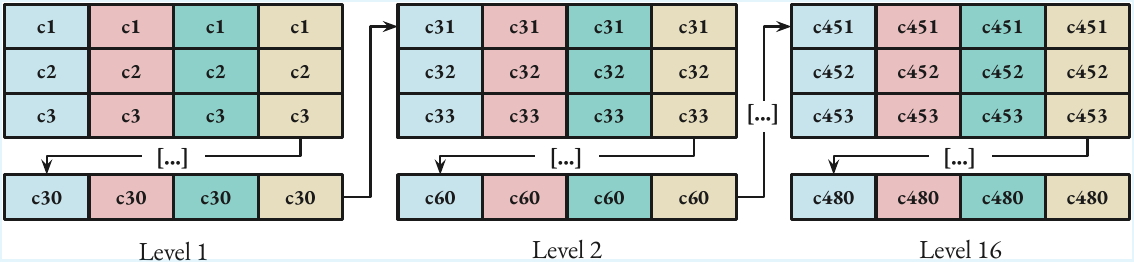}
    \caption{Batch 1 storage for IVFPQ on GIST.}\label{fig:storage}
    \vspace{-3mm}
  \end{figure}%
}

\newcommand{\figSpectrumShapingPQ}{%
  \begin{figure*}[t]
    \centering
    \includegraphics[width=\textwidth]{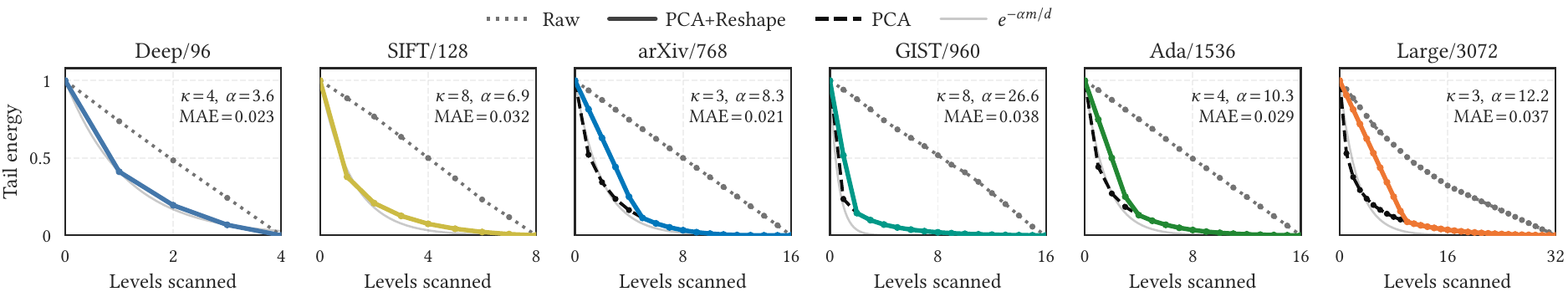}
    \caption{Per-dataset tail energy as a function of levels scanned for raw axes (dotted), PCA only (dashed), and \name's PCA+Reshape transform (solid). Each panel reports the shaping cap~$\kappa$ and the fitted PCA-only exponential decay rate~$\alpha$ used by the cost model. PCA gives the strongest compaction; PCA+Reshape trades some compaction for PQ-friendly level energies.}
    \label{fig:pq-shaping}
    \label{fig:energy-compaction}
    \label{fig:spectrum-shaping}
  \end{figure*}%
}

\newcommand{\figOrthogonalTransforms}{%
  \begin{figure*}[t]
    \centering
    \includegraphics[width=\textwidth]{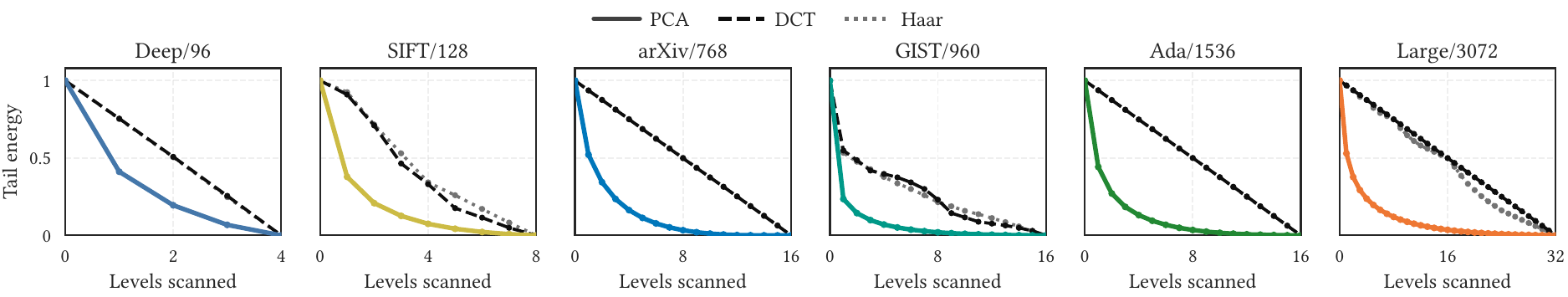}
    \caption{Per-dataset tail energy as a function of levels scanned for PCA, DCT, and Haar wavelet transforms.}
    \label{fig:orthogonal-transforms}
  \end{figure*}%
}

\newcommand{\figExhaustiveQps}{%
  \begin{figure}[t]
    \centering
    \includegraphics[width=\columnwidth]{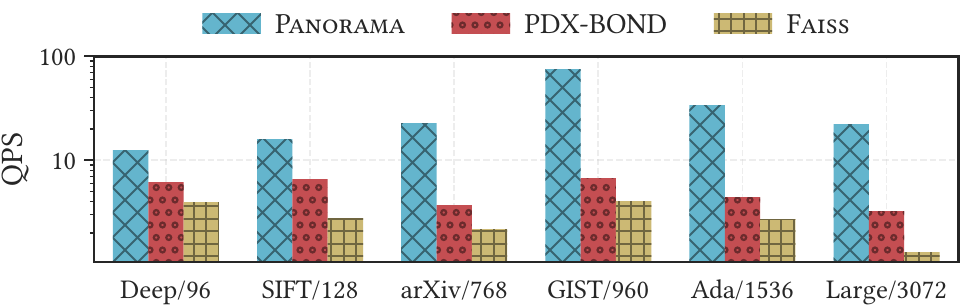}
    \caption{Throughput at exhaustive IVF probe ($\mathrm{nprobe}=\mathrm{nlist}$).}
    \label{fig:exhaustive-qps}
  \end{figure}%
}

\newcommand{\figMainWallCluster}{%
  \begin{figure*}[t]
    \centering
    \includegraphics[width=\textwidth]{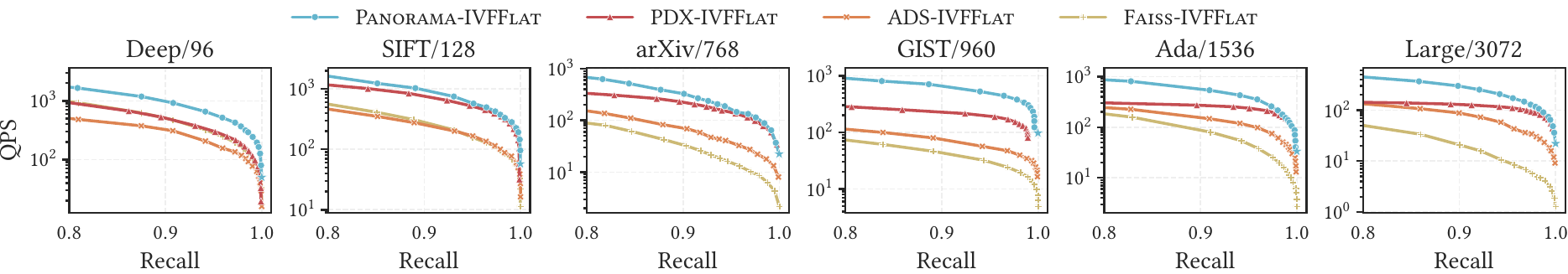}\\[0pt]
    \includegraphics[width=\textwidth]{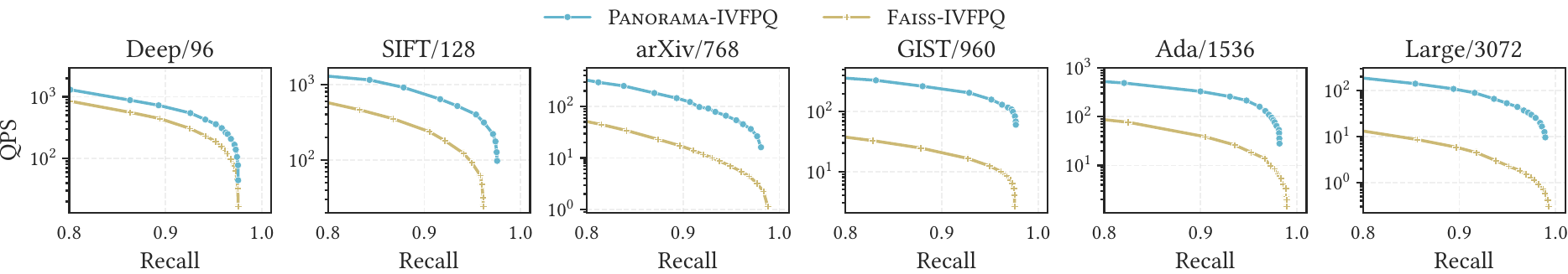}
    \caption{QPS vs.\ recall on cluster-based indexes. \textbf{Row 1:} \faiss-\ivfflat vs. PDX-\ivfflat vs. \name-\ivfflat. The best \name-\ivfflat QPS at full-recall (given by $\varepsilon{=}1$) by $\star$. \textbf{Row 2:} \faiss-\ivfpq vs. \name-\ivfpq.}
    \label{fig:main-wall-cluster}
  \end{figure*}%
}

\newcommand{\figMainWallGraph}{%
  \begin{figure*}[t]
    \centering
    \includegraphics[width=\textwidth]{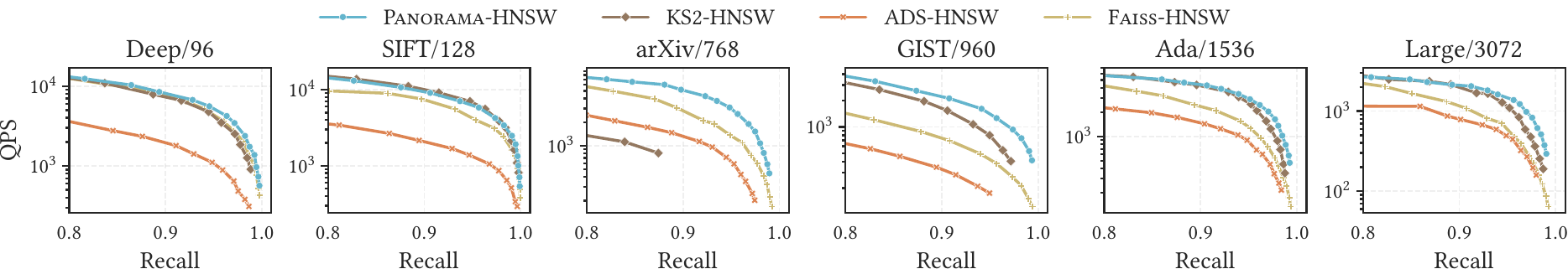}\\[0pt]
    \includegraphics[width=\textwidth]{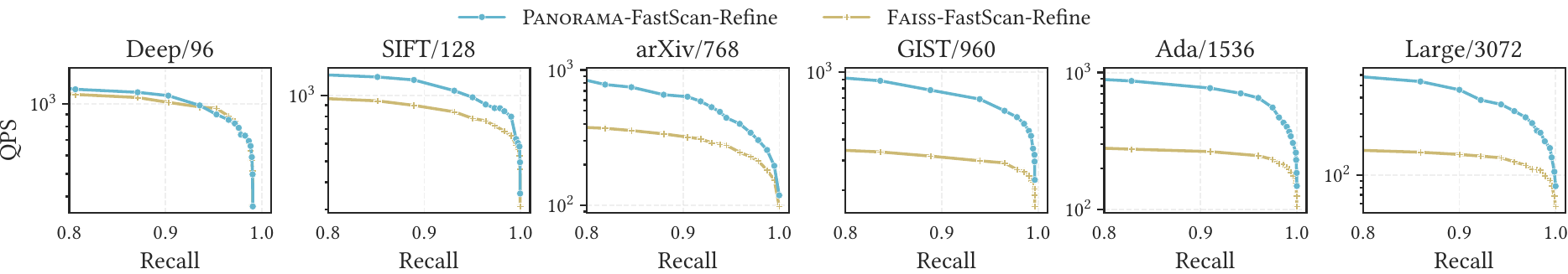}
    \caption{QPS vs.\ recall on graph- and refine-based indexes. \textbf{Row 1:} \faiss-\hnsw vs. KS2-\hnsw vs. ADSampling-\hnsw vs. \name-\hnsw. \textbf{Row 2:} \faiss-FastScan-Refine vs. \name-FastScan-Refine.}
    \label{fig:main-wall-graph}
  \end{figure*}%
}

\newcommand{\figMainWall}{\figMainWallCluster\figMainWallGraph}

\newcommand{\figTransformContribution}{%
  \begin{figure}[t]
    \centering
    \includegraphics[width=\columnwidth]{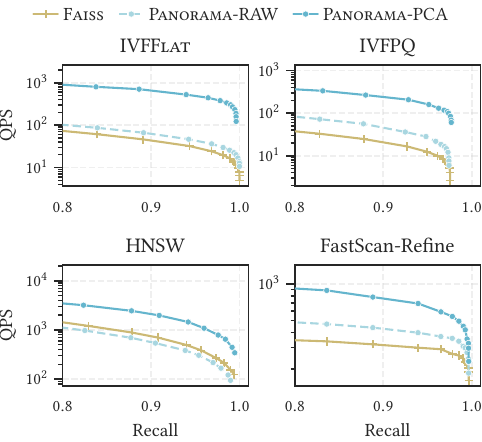}
    \caption{Transform contribution on GIST.}
    \label{fig:org-vs-trans}
  \end{figure}%
}

\newcommand{\figSystemsContribution}{%
  \begin{figure}[t]
    \centering
    \includegraphics[width=\columnwidth]{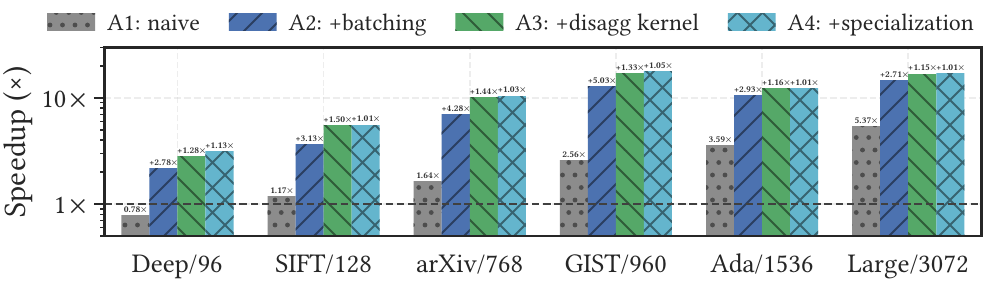}
    \caption{Systems contribution on GIST on IVF-PANO.}
    \label{fig:systems-contribution}
  \end{figure}%
}

\newcommand{\figExhaustiveEpsKappa}{%
  \begin{figure}[t]
    \centering
    \includegraphics[width=\columnwidth]{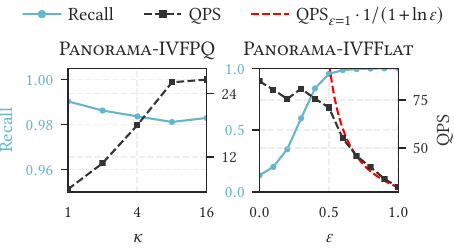}
    \caption{Pruning-knob ablation on Ada.}
    \label{fig:exhaustive-eps-kappa}
  \end{figure}%
}

\newcommand{\figHyperAblation}{%
  \begin{figure}[t]
    \centering
    \includegraphics[width=\columnwidth]{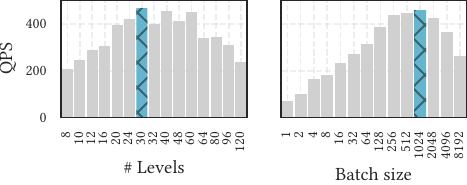}\hfill
    \caption{Second-order hyperparameter ablations on GIST.}
    \label{fig:hyper-ablation}
  \end{figure}%
}

\newcommand{\figOOD}{%
  \begin{figure}[t]
    \centering
    \includegraphics[width=\columnwidth]{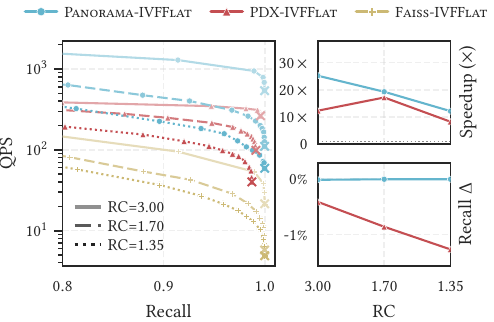}
    \caption{Out-of-distribution query robustness on GIST.}
    \label{fig:ood-rc}
  \end{figure}%
}

\newcommand{\figSemiEmpirical}{%
  \begin{figure}[t]
    \centering
    \includegraphics[width=\columnwidth]{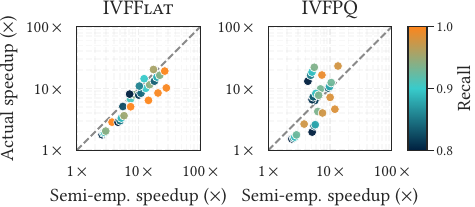}
    \caption{Measured vs.\ predicted end-to-end speedups on IVF and IVFPQ.}
    \label{fig:semi-empirical}
  \end{figure}%
}

\begin{document}
\title{\name: Fast-Track Nearest Neighbors}

\author{Alexis Schlomer}
\authornotemark[1]
\affiliation{%
  \institution{Carnegie Mellon University}
}
\email{aschlome@cs.cmu.edu}

\author{Akash K. Nayar}
\authornotemark[1]
\affiliation{%
  \institution{Carnegie Mellon University}
}
\email{akashnay@cs.cmu.edu}

\author{Vansh Ramani}
\authornote{Denotes equal contribution.}
\affiliation{%
  \institution{IIT Delhi \& CMU \& UCPH}
}
\email{cs5230804@cse.iitd.ac.in}

\author{Sayan Ranu}
\affiliation{%
  \institution{Indian Institute of Technology Delhi}
}
\email{sayanranu@cse.iitd.ac.in}

\author{Jignesh M. Patel}
\affiliation{%
  \institution{Carnegie Mellon University}
}
\email{jigneshp@cs.cmu.edu}

\author{Panagiotis Karras}
\affiliation{%
  \institution{University of Copenhagen}
}
\email{piekarras@gmail.com}
\begin{abstract}
\alexis{Candidate \emph{verification} is the primary bottleneck in Approximate Nearest-Neighbor Search (\anns) pipelines for high-dimensional neural embeddings. We introduce \name, a state-of-the-art refinement technique that accelerates verification by exploiting the inherent spectral decay of these embeddings. Using PCA to compact signal energy, \name evaluates candidate distances incrementally. At each step, it computes a strict lower bound on the full-vector distance, dynamically pruning candidates the moment this bound exceeds the running $k$-th nearest neighbor distance. While PCA's concentration of variance facilitates pruning, it breaks the uniform-variance assumption required by Product Quantization (PQ). To resolve this, we introduce a variance-shaping step that redistributes energy across subvectors to render accretive refinement compatible with quantized indexes. Optimized for modern memory hierarchies via vectorized bulk-pruning and cache-conscious data layouts, \name has been upstreamed into the \faiss library across major index families (IVFPQ/Flat, \hnsw, and Refine). \name achieves higher QPS at any target recall with a cost that provably scales inversely with dataset spectral decay, delivering end-to-end speedups of up to \speedup~and outperforming probabilistic methods across all recall bands.}
\end{abstract}
\maketitle


\section{Introduction}

\pkv{The proliferation of large-scale neural embeddings has transformed machine learning applications, from computer vision and recommendation systems~\citep{lowe2004distinctive, koren2009matrix} to bioinformatics~\citep{altschul1990basic} and retrieval-augmented generation (RAG) pipelines~\citep{lewis2020retrieval, gao2023retrieval}. As embedding models produce vectors with thousands of dimensions---OpenAI's $\mathtt{text\text{-}embedding\text{-}3\text{-}large}$~\citep{neelakantan2022textcodeembeddingscontrastive} reaches~3072---the need for real-time Approximate Nearest-Neighbor Search (\anns) intensifies.}


\paragraph{The refinement bottleneck} \pkv{Current \anns methods fall into four major families: \emph{graph-based} (HNSW~\citep{hnsw}, DiskANN~\citep{diskann}), \emph{clustering-and-quantization-based} (IVFPQ~\citep{pq, ivf}, ScaNN~\citep{scann}), \emph{tree-based} (kd-trees~\citep{kdtree_original}, FLANN~\citep{flann}), and \emph{hash-based} (LSH~\citep{lsh_original,e2lsh}, multi-probe LSH~\citep{multiprobe_lsh}). All families share a pipeline of two phases~\citep{efficient_indexing}: \emph{filtering} reduces the database to a set of candidates~\candidates and \emph{refinement} (verification) selects~$k$ candidates nearest to the query. Prior work in \anns has almost exclusively targeted the filtering phase, assuming that refinement is inconsequential. This assumption holds on low-dimensional embeddings; on modern neural embeddings, however, refinement consumes most of end-to-end query latency, with the share rising monotonically in dimensionality~\citep{adsampling}.}

\paragraph{Why existing remedies underperform} \pkv{Prior work addresses the refinement bottleneck primarily through \emph{probabilistic} distance estimation: ADSampling~\citep{adsampling} randomly subsamples coordinates, while FINGER and KS2~\citep{finger, pkf} approximate inner products via angle estimation. Consequently, these methods hit a structural recall ceiling and cannot guarantee exact k-NN regardless of the time budget. Furthermore, they are poorly suited for cluster-based indexes that utilize sequential scans. PDX~\citep{pdx} mitigates the memory bottleneck via a \akash{columnar} layout, yet suffers this same recall deficiency or, in its deterministic variant PDX-BOND, low throughput. Finally, none of these approaches composes with \emph{product quantization} (PQ)~\citep{pq}, a ubiquitous compression primitive in billion-scale \anns. This paper introduces \name, a deterministic and hardware-friendly refinement method that eliminates this compromise. By providing a mathematically principled pruning bound, \name delivers superior throughput at \emph{every} target recall band—up to and including 100\% recall—while natively integrating with PQ and other existing index families.}

\paragraph{Contributions} \pkv{Our contributions are as follows:}

\begin{itemize}[leftmargin=*,itemsep=1pt,topsep=2pt]
\item \textbf{Accretive refinement.} \pkv{\name accumulates~$L_2$ distance terms incrementally along the PCA basis, maintains tight Cauchy--Schwarz lower bounds on the fly, and prunes any candidate whose lower bound exceeds the running~$k^\mathrm{th}$-nearest-neighbor threshold. This pruning is controlled by a tunable parameter $\varepsilon$, exact at~$\varepsilon=1$, and relaxed for the sake of efficiency with~$\varepsilon<1$. This scheme fits into graph, tree, clustering, and hash-based indexes without changing their filtering phase.}

\item \textbf{Theoretical grounding.} \pkv{We show that the expected verification cost scales inversely with an exponential PCA decay factor observed in real-world embeddings (Proposition~\ref{thm:final_complexity}) and generalize this guarantee to out-of-distribution queries (Proposition~\ref{thm:ood_complexity}).}

\item \textbf{Compatibility with product quantization.} \pkv{While PCA's aggressive variance concentration aids pruning, it destabilizes downstream product quantization (PQ) and erodes recall. To counteract this, we apply an orthogonal transformation~$R$ that flattens coordinate variances to a target diagonal~$t \prec \lambda$ while preserving covariance eigenvalues~$\lambda$. We construct~$R$ by a Robin Hood sequence of Haar-random rotations that transfers surplus variance across PQ blocks, followed by an intra-block Haar-random rotation to reduce residual anisotropy. Together, these steps preserve PCA's pruning efficacy while ensuring PQ compatibility.}

\item \textbf{Systems co-design and integration.} \pkv{We co-design \name with the memory hierarchy, providing a \emph{batched} variant for contiguous layouts, using SIMD bulk-pruning kernels over a level-major data layout, and a \emph{no-batch} (point-centric) variant for non-contiguous layouts. \name is available in \faiss across HNSW, IVFPQ, \ivfflat, and a PQ-refine step (FastScan\akash{-Refine}), delivering up to \speedup~end-to-end speedups.}
\end{itemize}

\section{Related Work}




\pkv{Approximate Nearest-Neighbor Search (ANNS) pipelines generally operate in two phases: a filtering phase that retrieves a candidate set, and a refinement phase that computes exact distances to find the true top-$k$ neighbors. Historically, the literature has heavily optimized the filtering phase. \emph{Graph-based methods}~\citep{nsw, hnsw, diskann}, \emph{tree-based partitions}~\citep{kdtree_original, flann}, and \emph{Locality-Sensitive Hashing} (LSH)~\citep{lsh_original, multiprobe_lsh} all employ distinct spatial navigation strategies to reduce the candidate pool. However, as embedding dimensionalities grow, the exact distance evaluations required at the leaves, neighborhoods, or hash buckets of these structures become the dominant computational bottleneck. \name is designed to operate atop any of these filtering structures to accelerate this final distance computation.}

\alexis{\emph{Clustering-and-quantization methods} partition space and compress vectors to reduce storage and accelerate search. IVF~\cite{ivf} builds inverted indexes via $k$-means clustering to map queries to nearby centroids. IVFPQ~\cite{pq} applies \emph{product quantization} (PQ) within clusters, decomposing vectors into independently quantized subvectors. Advanced variants further minimize quantization error via learned rotations (OPQ~\cite{opq}), multi-cluster assignments (Multi-D-ADC~\cite{babenko2015}), anisotropic quantization (ScaNN~\cite{scann}), or by introducing theoretical error bounds for extreme quantization (RaBitQ~\citep{rabitq}).}

\alexis{While these methods excel at compression, scanning encoded vectors remains a computational bottleneck, as traditional implementations require exhaustive distance calculations across all subvectors. Recent approaches address this limitation by optimizing both storage and scanning. Variance-Aware Quantization (VAQ)~\citep{vaq}, for instance, abandons PQ's uniform bit-budget to adapt dictionary sizes proportionally to each subspace's explained variance. Crucially, VAQ bypasses exhaustive scanning by employing triangle inequality bounds and early-abandoning heuristics to mathematically prune candidates on the fly. While VAQ addresses variance skew by heavily modifying the underlying quantizer, \name does the exact opposite: we introduce an orthogonal spectrum-shaping transformation that flattens the data's variance. This ensures strict compatibility with highly-optimized, uniform PQ implementations while preserving enough front-loaded energy to power accretive early-abandoning.}

\pkv{Beyond quantization, dedicated \emph{refinement techniques} attempt to accelerate the final distance computations directly. Probabilistic methods such as ADSampling~\citep{adsampling}, PDX~\citep{pdx}, and FINGER~\citep{finger} rely on stochastic sub-sampling or proxy metrics, which intrinsically degrade maximum recall. Conversely, exact refinement methods, including BOND~\citep{bond} and Stepwise~\citep{stepwise}, establish strict distance bounds to prevent false dismissals but suffer from loose bounds. By forgoing probabilistic estimation in favor of mathematically rigorous Cauchy-Schwarz bounds along a shaped PCA basis, \name generates tight pruning bounds, allowing it to exceed the throughput of probabilistic methods across all recall bands.}

\pkv{Finally, some works refer to refinement over~$k' = k_{\mathrm{factor}} \cdot k$ over-fetched candidates as \emph{`reranking'} (e.g., using coarse indices like IVFPQFastScan~\citep{fastscan} for rapid initial retrieval followed by unquantized $k$NN). Because \name accelerates the underlying mathematical operations of distance computation, it serves as a drop-in replacement to accelerate this exact reranking stage as well.}

\section{Panorama: Distance Computation}\label{sec:panorama}

\begin{problem}[$k$NN refinement]\label{prob:kNN}
\pk{Given a query vector~$\query \in \mathbb{R}^d$ and a candidate set~$\rev{\candidates} = \{ \xvec_1, \ldots, \xvec_{\rev{N}}\}$, find the set~$\mathcal{S} \subseteq \rev{\candidates}$ such that~$|\mathcal{S}| = k$ and~$\forall \svec \in \mathcal{S}, \xvec \in \rev{\candidates} \setminus \mathcal{S}: \|\query - \svec\|_2 \leq \|\query - \xvec\|_2$.}
\end{problem}

\begin{problem}[ANN index]\label{prob:index}
\pk{An \emph{approximate nearest neighbor} index is a function~$\mathcal{I}: \mathbb{R}^d \times \mathbb{D} \to 2^{|\mathbb{D}|}$ that maps a query $\query$ and a database~$\mathbb{D}$ to a candidate set~$\rev{\candidates} = \mathcal{I}(\query, \mathbb{D})$, where~$\rev{\candidates}$ contains the true~$k$-nearest neighbors with high probability.}\footnote{\pk{Some indexes like HNSW interleave filtering and refinement; we treat the exact-distance computations on visited nodes as the refinement phase.}}
\end{problem}

\pk{\Cref{prob:kNN} is the computational bottleneck: with~$N$ candidates, naive refinement computes~$\|\query - \xvec_i\|_2^2 = \sum_{j=1}^d (\query_j - \xvec_{i,j})^2$ for each candidate, requiring~$\Theta(N \cdot d)$ operations.}

\paragraph{Accretive bounds.} \pkv{We decompose the squared Euclidean distance through a norm-preserving orthogonal transform~$T:\mathbb{R}^d \to \mathbb{R}^d$, following~\citet{stepwise}:}

\begin{equation}\label{eq:distance_decomp}
\|\query - \xvec\|^2 = \|T(\query)\|^2 + \|T(\xvec)\|^2 - 2\langle T(\query), T(\xvec)\rangle.
\end{equation}

\pk{Given thresholds~\(0 = m_0 < m_1 < \cdots < m_L = d\) that partition the dimensions into \(L\) levels, we define partial inner products and tail (residual) energies:}

\begin{align}\label{eq:partial_products}
p^{(\ell_1, \ell_2)}(\query, \xvec) &=\!\!\! \sumell \!\! T(\query)_j T(\xvec)_j, \nonumber\\
R_{T(\query)}^{(\ell_1, \ell_2)} &=\!\!\! \sumell \!\! T(\query)_j^2, \quad
R_{T(\xvec)}^{(\ell_1,\ell_2)} =\!\!\! \sumell \!\! T(\xvec)_j^2.
\end{align}

\pk{The Cauchy--Schwarz inequality~\citep{horn2012matrix} yields a monotonically tightening lower bound on the squared distance:}

\begin{equation}\label{eq:lower_bound}
\begin{split}
\lb^\ell(\query, \xvec) = R_{T(\query)}^{(0, d)} + R_{T(\xvec)}^{(0, d)} \\
- 2 \!\left( p^{(0,\ell)}(\query, \xvec) + \sqrt{R_{T(\query)}^{(\ell,d)} R_{T(\xvec)} ^{(\ell,d)}} \right) \leq  \|\query - \xvec\|^2.
\end{split}
\end{equation}

\pkv{We further introduce a \emph{pruning aggressiveness} parameter~$\varepsilon \in [0,1]$. At level~$\ell$, the distance comprises an exact accumulated term minus the unknown tail inner product. Let~$A^\ell = R_{T(\query)}^{(0,d)} + R_{T(\xvec)}^{(0,d)} - 2p^{(0,\ell)}(\query,\xvec)$ and~$C^\ell = \sqrt{ R_{ T(\query)} ^{ (\ell, d)} R_{ T( \xvec ) }^{(\ell,d)} }$. \vansh{The strict Cauchy--Schwarz bound subtracts the worst-case tail term~$2C^\ell$; reducing this correction gives the relaxed score:}}
\begin{equation}\label{eq:epsilon_bound}
\lb_\varepsilon^\ell(\query,\xvec) = A^\ell - 2\varepsilon C^\ell.
\end{equation}

\paragraph{Soundness at \boldmath$\varepsilon=1$.} \alexis{Setting $\varepsilon = 1$ recovers the exact Cauchy--Schwarz lower bound (Equation~\eqref{eq:lower_bound}), \vansh{so pruning cannot remove any candidate that could still enter the top-$k$ over the candidate set.} Lowering~$\varepsilon$ relaxes this worst-case tail correction to enable more aggressive pruning and higher throughput, trading off \vansh{strict exactness for a controlled recall--throughput knob}. As~\Cref{fig:main-wall-cluster,fig:main-wall-graph} show, \vansh{this relaxation consistently yields superior performance at any recall target while the strict mode remains available whenever exact refinement is needed.}}
\begin{algorithm}[H]
\caption{\name: Iterative Distance Refinement}\label{alg:panorama_all}
\begin{algorithmic}[1]
\scriptsize
\State \textbf{Input:} Query $\mathbf{q}$, candidate set $\mathcal{C} = \{\mathbf{x}_1, \ldots, \mathbf{x}_{N'}\}$, transform $T$, levels $m_1 < \cdots < m_L$, $k$, batch size $B$, pruning parameter $\varepsilon \in [0,1]$
\State \textbf{Precompute:} $T(\mathbf{q})$, $\|T(\mathbf{q})\|^2$, and tail energies $R_q^{(\ell,d)}$ for all $\ell$
\State \textbf{Initialize:} Global exact distance heap~$H$ (size $k$), global threshold~$\kub \leftarrow +\infty$
\State Compute exact distances of first~$k$ candidates, initialize $H$ and $\kub$ \label{lin:initial}
\For{each batch $\mathcal{B} \subset \mathcal{C}$ of size $B$}
    \For{$\ell = 1$ to~$L$}
        \For{each candidate $\xvec \in \mathcal{B}$}
            \If{$\lb_\varepsilon^\ell(\query,\xvec)>\kub$}\label{lin:prune} \Comment{$\varepsilon=1$ is exact; smaller values prune more}
                \State Mark $\xvec$ as pruned \label{lin:break} \Comment{If threshold exceeded, prune candidate}
                \State \textbf{continue}
            \EndIf
        \EndFor
    \EndFor
    \For{each unpruned candidate $\xvec \in \mathcal{B}$}
        \State Push~$(\lb^L(\query, \xvec), \xvec)$ to~$H$ as exact entry \label{lin:push} \Comment{$\lb^{L}(\query, \xvec)$ is ED as $\ell = L$}
        \If{$d < \kub$}
            \State Update~$\kub = \kth$ distance in~$H$; Crop~$H$ \label{line:heapupdate}
        \EndIf
    \EndFor
\EndFor
\State \textbf{return} Candidates in~$H$ (top~$k$ with possible ties at~$\kth$ position)
\end{algorithmic}
\end{algorithm}

\paragraph{Algorithm.} \pkv{\Cref{alg:panorama_all} is the \name\ refinement loop. It maintains a heap~$H$ of exact $k$NN distances seen so far, initialized on the first~$k$ candidates, and the associated~\kth smallest distance~\kub. For each subsequent batch, it advances all candidates one level at a time, prunes any with lower bound~$\lb_\varepsilon^\ell(\query, \xvec) > \kub$ (Line~\ref{lin:prune}), and bulk-commits the survivors' exact distances to~$H$ at the final level (Lines~\ref{lin:push}--\ref{line:heapupdate}). 
As memory layouts across state-of-the-art indexes vary wildly, we decouple the pruning strategy from the layout via the \emph{batch size} parameter~$B$:
}




\begin{table}[h]
\centering
\panoramaTableStyle
\caption{\pkv{\name's two modes, driven by batch size.}}\label{tab:variants}
\begin{tabular}{@{}lcl@{}}
\toprule
\textbf{Variant} & \boldmath$B$ & \textbf{Applicable indexes} \\
\midrule
Batched (clustered)           & $B > 1$ & \ivfflat, IVFPQ \\
No-batch (point-centric) & $B = 1$ & HNSW, Refine \\
\bottomrule
\end{tabular}
\end{table}

\pkv{When the index organizes candidates contiguously (IVFFlat, IVFPQ), we set~$B$ to~1024 to balance systems affordances with pruning power, as discussed in \Cref{subsec:sys_ivfflat}. When candidates arise non-contiguously (HNSW, Refine), we set~$B = 1$: the inner batch loop collapses and each candidate is early-abandoned at its first offending level, which is the only feasible mode under scattered memory access. \Cref{sec:integration} elaborates on the systems consequences.}

\pkv{Let~$\rho_i \in \{m_0, \ldots, m_L\}$ be the dimension at which candidate~$\xvec_i$ is pruned (or survives, with~$\rho_i = d$). The total cost is~$\sum_{i=1}^{N} \rho_i$, so with~$\phi = \frac{\mathbb{E}[\rho]}d$ the average fraction of dimensions processed, the expected cost is~$\bigO(\phi d N)$. We make this scaling precise in \Cref{sec:theory}.}

\pkv{\name relies on (i)~a transform~$T$ that concentrates energy in leading dimensions, enabling tight bounds---achieved through PCA (\Cref{sec:transforms}) 
and (ii)~level thresholds~$m_\ell$ that balance the per-level control overhead against pruning granularity.}

\figOrthogonalTransforms

\section{Transform and Variance Shaping}\label{sec:transforms}

\pkvall{The scheme of Section~\ref{sec:panorama} reduces the distance estimation per candidate~\xvec and level~$\ell$ to the geometric mean~$\sqrt{R_{ T(\query) }^{ (\ell, d) }\, R_{ T(\xvec) }^{ (\ell,d) }}$ of tail energies. This quantity drops for each~\xvec as~$\ell$ grows; when it falls below the running gap between \xvec's known distance and the current~\kth distance, \xvec is ruled out of the~$k$NN set, hence we stop reading its coordinates. The task of the orthogonal transform~$T$ is to make those trailing tail energies shrink promptly.

This section addresses the following two questions:

\begin{itemize}
\item \emph{Which} orthogonal transform shrinks the tail most effectively (\S\ref{ssec:pca})? We argue that PCA is the natural answer: a closed-form, data-adaptive transform that minimizes the expected tail energy at each level, and whose spectrum yields the decay rate.
\item \emph{How} do we reconcile PCA with product quantization (PQ), given that PCA breaks PQ's uniform bit-budget assumption~(\S\ref{ssec:shaping})? We amend PCA with a process that evens out energy between and within PQ blocks while retaining most of its front-loading.
\end{itemize}}

\subsection{PCA as the closed-form transform}\label{ssec:pca}

\pkv{Given a dataset~$\dataset \subset \mathbb{R}^d$ with empirical covariance~$\Sigma$, PCA (\emph{principal component analysis}) returns the orthogonal matrix~$T \in \mathbb{R}^{d\times d}$ whose rows are the eigenvectors of~$\Sigma$, ordered so that the corresponding eigenvalues satisfy~$\lambda_1 \ge \cdots \ge \lambda_d \ge 0$. As an orthogonal transform, PCA preserves Euclidean geometry: $\|T(\xvec)\|_2=\|\xvec\|_2$ and $\|T(\query)-T(\xvec)\|_2=\|\query-\xvec\|_2$, hence transformed vectors retain the distance decomposition of Equation~\eqref{eq:distance_decomp} and the lower bound of Equation~\eqref{eq:lower_bound} is \emph{lossless}. PCA is singled out among orthogonal transforms by its energy \emph{distribution} across coordinates.}

\figSpectrumShapingPQ

\paragraph{Why PCA is the right basis?} \pkv{Orthogonal transforms conserve energy: the total~$\sum_j \|T(\xvec)_j\|^2 = \|\xvec\|^2$ is fixed, so any reduction in tail energy arises by moving energy into leading coordinates. The movable amount is determined by the spectrum of~$\Sigma$ via a classical variational fact (Ky~Fan / Schur--Horn~\cite{horn2012matrix}): the maximum energy that an orthogonal transform can place in the first~$m$ coordinates is~$\sum_{j=1}^m \lambda_j$, attained when the leading rows of~$T$ span the top-$m$ eigenvalues of~$\Sigma$---which is exactly what PCA provides. PCA is therefore optimal at every level simultaneously, leaving the smallest possible expected tail~$\sum_{j>m}\lambda_j$.}

\paragraph{How about other orthogonal transforms?} \pkv{Two alternatives---the Discrete Cosine Transform~\cite{dct_smooth} and the Haar wavelet basis~\cite{mallat1999wavelet}---also concentrate energy into leading coordinates, but only provided that the data has strong \emph{local self-similarity}, i.e., adjacent coordinates of correlated values, as in smooth signals or natural images. These \emph{fixed} (non-data-adaptive) bases succeed when the data already aligns with them. In unstructured embedding spaces such as text representations and learned features, no such locality exists, hence these fixed bases do not improve on the original axes. By contrast, PCA's eigenvectors are by definition the directions of maximum variance \emph{in the data at hand}, hence PCA adapts to the data's covariance structure. As \Cref{fig:orthogonal-transforms} illustrates, PCA compacts energy more than DCT and Haar wavelet transforms, regardless of the underlying data.}

\paragraph{Energy compaction and the parameter~$\alpha$.} \pkv{While PCA is the right basis, the way the tail shrinks is an empirical property of a dataset. We define the expected residual energy at level~$m$ as the normalized PCA tail variance,
\[
E(m) = \frac{\sum_{j=m+1}^{d}\lambda_j}{\sum_{j=1}^{d}\lambda_j},
\]
i.e., the fraction of total variance retained by the trailing $d-m$ coordinates. Across all six datasets in our study, $E(m)$ is modeled well by the exponential:
\[
E(m) \;\approx\; e^{-\alpha \frac md},
\] 
with a data-specific constant~$\alpha>0$ obtained from the empirical PCA spectrum. Figure~\ref{fig:energy-compaction} reports the empirical~$E(m)$ and the fitted exponential on each dataset; confirming that the exponential model is faithful to the data. This shape is what classical spectral theory predicts for analytic covariance kernels---smooth neural embeddings (GELU, softmax) and Gaussian-family kernels among them; we develop the connection in Section~\ref{sec:theory}. Here, $\alpha$ is a measured number per dataset. Larger~$\alpha$ denotes that the tail collapses more rapidly, hence the Cauchy--Schwarz remainder shrinks, and \name prunes earlier. We concretize the connection in Section~\ref{sec:theory}, showing that the expected refinement cost scales as~$\frac{Nd}\alpha$.}

\subsection{Spectrum shaping for PQ-compatibility}\label{ssec:shaping}

\pkv{PCA is excellent for pruning alone, as it aggressively concentrates variance into the leading dimensions. However, in the IVFPQ index, \name has to coexist with product quantization~(PQ), which clashes with that concentration. We recall how PQ works, explain why PCA breaks it, and describe the two-stage procedure that fixes the problem while preserving most of PCA's compaction.}

\paragraph{What PQ expects} \pkv{Product quantization~\cite{pq} splits each $d$\hyp{}dimensional vector into~$M$ contiguous subvectors of size~$\sfrac dM$ and quantizes each subvector with its own $2^{n_\mathrm{bits}}$-entry codebook learned by~$k$-means. Each subvector receives the same number of code bits, under the implicit assumption that variance is even across subvectors. If one subvector carries almost all the energy, an equal-bit codebook leads to underrepresentation of the data; if another subvector is almost constant, the codebook budget is wasted on noise.}

\paragraph{Why raw PCA breaks PQ} \pkv{The concentration of energy in the first dimensions, which confers to PCA a pruning advantage, clashes with PQ. As the leading subvectors carry nearly all the variance and the trailing ones collapse to near-constant, PQ underserves the informative leading blocks while wasting a fixed-bit budget on near-zero blocks. To use \name with IVFPQ, we should reconcile the PCA per-level energies with PQ, while retaining most of the front-loading that facilitates pruning. We do so in two steps: a coarse \emph{inter-level} redistribution of energy among levels (i.e., blocks), followed by a fine \emph{intra-level} isotropization that spreads each level's energy over its dimensions.}

\paragraph{Procedure 1: Robin Hood inter-level shaping} \pkv{We split the~$d$ PCA coordinates into~$L$ contiguous levels of size~$b = \sfrac dL$, which we use to refine bounds. Let~$E_\ell = \sum_{j=(\ell-1)b+1}^{\ell b}\lambda_j$ be the energy of level~$\ell$. PCA endows the leading levels more than the trailing ones, $E_1 \gg E_L$; PQ requires something closer to a flat profile. A user-chosen hyperparameter~$\kappa \ge 1$ specifies a cap~$\tau = \kappa\bar{E}$ on how much energy a level may carry, where~$\bar{E} = \sfrac{\sum_\ell E_\ell}L$ is the average; $\kappa=1$ forces perfect equalization; larger~$\kappa$ allows leading levels to remain richer. We sweep levels from front to back, leaving any level with $E_\ell \le \tau$ unchanged. At a level~$\ell$ that violates the cap, $E_\ell>\tau$, we apply a Haar-random orthogonal rotation to its dimensions along with a minimal suffix of subsequent dimensions---so that the rotated level's expected energy drops to~$\tau$. As the rotation is orthogonal, it preserves total energy, redistributing the surplus. This is the \emph{Robin Hood} metaphor---a rich leading level spills into poorer trailing ones to meet the cap. The procedure terminates once every level satisfies~$\tau$, and the resulting matrix~$R$ is composed with the PCA basis to yield a single transform, $R \cdot T_{\mathrm{PCA}}$.}

\paragraph{Procedure 2: Intra-level isotropization} \pkv{Following Procedure~1, each level's energy is at most~$\tau$. However, the leading coordinates \emph{within} a level may still hold most of the energy, and thus, as~$M > L$ (i.e., there are more PQ subvectors than levels), cause imbalance for PQ. To eliminate this residual anisotropy, we apply an independent Haar-random orthogonal transform~$U_\ell\in\mathrm{O}(b)$ within each level's dimensions. As~$U_\ell$ is orthogonal, it preserves each level's total energy and spreads it uniformly in expectation across~$b$ dimensions, as $k$-means codebooks require.}

\paragraph{The combined transform} \pkv{The complete transform is the composition~$U \cdot R \cdot T_{\mathrm{PCA}}$, where~$T_{\mathrm{PCA}}$ is the PCA basis, $R$ the Robin Hood rotation of Procedure~1, and~$U$ the block-diagonal isotropization of Procedure~2. The composition is orthogonal, preserving soundness, while the per-vector storage and runtime are identical to vanilla PCA. As a side benefit, the intra-level Haar rotation by~$U$ makes the residual coordinates isotropic, which the analysis of Section~\ref{sec:theory} relies on for the~$\varepsilon$-relaxation result of Proposition~\ref{thm:eps_speedup}.}

\pkv{Figure~\ref{fig:spectrum-shaping} reports the per-level tail energy on each of our six datasets for three transforms: raw (no rotation), PCA-only, and PCA+Shaping (the full shaped transform with the shown hyperparameter~$\kappa$). PCA-only attains the strongest pruning and determines the compaction coefficient~$\alpha$ used in our cost model (\Cref{sec:theory}). PCA+Shaping forfeits some compaction to attain level energies that fit within the cap~$\tau$, for the sake of IVFPQ; the gap between the two curves is small, as Procedure~1 only applies when a level exceeds the cap. The resulting transform is simultaneously amenable to accretive pruning \emph{and} product quantization---allowing \name to work with IVFPQ while maintaining high recall.}

\section{The Cost Model}\label{sec:theory}

\vansh{Section~\ref{sec:transforms} surfaced a single empirical quantity---the PCA tail-decay rate~$\alpha$---that characterizes how quickly the residual energy of an embedding collapses under the PCA basis. This section turns~$\alpha$ into a quantitative model of \name's refinement cost. We first justify PCA as the energy-optimal basis (\S\ref{ssec:basis}), then derive the core scaling law $\mathbb{E}[\mathrm{Cost}] \lesssim Nd/\alpha$ (\S\ref{ssec:cost-model}), and finally extend the bound to out-of-distribution queries (\S\ref{ssec:ood}) and to the speed--recall knob~$\varepsilon$ (\S\ref{ssec:eps}). Each result is paired with a direct empirical comparison; full proofs appear in Appendix~\ref{app:full-proofs}.}
\vansh{We use $f\lesssim g$ to denote upper-bound scaling after the finite-sample constants in the model have been made explicit.}

\subsection{Choice of Basis}\label{ssec:basis}

\vansh{\S\ref{ssec:pca} established that PCA is the energy-optimal orthogonal basis: no transform leaves less expected energy $\sum_{j>m}\lambda_j$ in the trailing coordinates at any level~$m$. How fast that tail shrinks is, however, a property of each dataset. Spectral theory provides a strong prior: analytic covariance kernels on a compact domain produce geometrically-decaying Mercer eigenvalues~\cite{little1984eigenvalues}, and smooth-activation neural embeddings inherit this through their tangent kernel~\cite{bietti2019inductive}. Figure~\ref{fig:energy-compaction} confirms it empirically---across all six benchmarks, the PCA tail collapses as $e^{-\alpha m/d}$, with mean absolute fit error below $0.04$. The fitted~$\alpha$ is the single empirical input to everything that follows.}

\subsection{Expected Refinement Cost}\label{ssec:cost-model}

\vansh{\name prunes a candidate at level~$m$ when its level-$m$ lower bound exceeds the running $k$-th best distance~$d_k$. Equivalently, pruning fires when the gap between the true squared distance and the lower bound drops below the \emph{pruning margin}}
\vansh{
\[
\margin(\xvec) \;:=\; \|\query - \xvec\|^2 - d_k.
\]
}
\vansh{True $k$-nearest neighbors satisfy $\margin \le 0$ by definition and are never pruned. Non-nearest candidates have positive margin in expectation: in high dimensions, query--database distances tightly concentrate around a mean~$\mu$, while $d_k$ tracks the lower tail of distances seen so far. A Dvoretzky--Kiefer--Wolfowitz argument~\cite{massart1990tight} on the empirical CDF places~$d_k$ a slowly-growing number of standard deviations below~$\mu$, so a typical non-nearest candidate has $\margin > 0$ with high probability.}

\vansh{Combining Cauchy--Schwarz with the exponential tail of \S\ref{ssec:basis} and a bounded-norm condition $\|T(\xvec)\| \le R$ (giving the prefactor $C_0 := 4R^2$), the level-$m$ gap satisfies}
\vansh{
\[
\|\query - \xvec\|^2 - \lb^m(\query, \xvec) \;\le\; C_0\, e^{-\alpha m / d}.
\]
}
\vansh{Pruning fires at the smallest~$m$ for which this gap drops below~$\margin$, giving capped per-candidate depth $\rho \approx \min\{d,(d/\alpha)[\ln(C_0/\margin)]_+\}$. Summing $\rho$ over $N$ candidates and absorbing the averaged margin terms into a finite-sample constant~$c$ yields the central result of the section.}

\begin{proposition}\label{thm:final_complexity}
\vansh{Under exponential PCA tail decay, bounded vector norms, and concentrated query-distance distributions, the expected number of coordinate evaluations \name performs over $N$ candidates satisfies}
\vansh{
\[
\mathbb{E}[\mathrm{Cost}] \;\lesssim\; \frac{c \cdot Nd}{\alpha}.
\]
}
\end{proposition}

\noindent\vansh{Here, $c$ is a finite-sample margin constant that absorbs the averaged per-candidate margin terms.}

\paragraph{Proof sketch.} \vansh{The argument has three pieces. First, we turn the energy-decay rate~$\alpha$ into a per-candidate pruning depth bound. Second, we model the pruning margin~$\margin$ through the order-statistic structure of the running threshold~$d_k$. Third, we sum the capped per-candidate depths over the~$N$ candidates and absorb the averaged margin terms into the finite-sample constant~$c$.}

\vansh{\emph{(a) Pruning depth.} Combining the Cauchy--Schwarz inequality with the exponential PCA tail and the bounded-norm condition, the gap between the true squared distance and the level-$m$ lower bound is:}
\vansh{
\[
\begin{aligned}
\|\query - \xvec\|^2 - \lb^m(\query, \xvec)
  &\;\le\; 4\sqrt{R_{T(\query)}^{(m,d)} R_{T(\xvec)}^{(m,d)}} \\
  &\;\approx\; 4\|\query\|\|\xvec\|\, e^{-\alpha m/d}
  \;\le\; C_0\, e^{-\alpha m/d},
\end{aligned}
\]
}
\vansh{where the first step is Cauchy--Schwarz, the second applies the exponential tail to query and database, and the third uses the bounded-norm condition along with the definition $C_0 := 4R^2$. Pruning occurs when this gap drops below the margin~$\margin$, i.e., at the smallest~$m$ with $C_0\, e^{-\alpha m/d} \le \margin$. Solving for~$m$ gives the per-candidate pruning depth}
\vansh{
\[
\rho \;\approx\; \min\!\left\{d,\ \frac{d}{\alpha}\Big[\ln\!\left(\frac{C_0}{\margin}\right)\Big]_+\right\}.
\]
}
\vansh{This depth is proportional to~$d/\alpha$ and logarithmic in the margin.}

\vansh{\emph{(b) Margin from a Gaussian order statistic.} The threshold~$d_k$ is, by construction, the $k$-th order statistic among processed squared distances. Intuitively, after processing $i \gg k$ candidates, $d_k$ tracks the lower tail of the distance distribution: if distances cluster around mean~$\mu$, then $d_k$ sits a (slowly growing) number of standard deviations \emph{below}~$\mu$ and stays below~$\mu$. A typical non-nearest candidate has true squared distance close to~$\mu$, so its margin $\margin = \|\query-\xvec\|^2 - d_k$ is typically positive. Formally, under the Gaussian distance approximation, the Dvoretzky--Kiefer--Wolfowitz (DKW) inequality~\cite{massart1990tight} on the empirical distance CDF bounds the threshold by $d_k \le \mu + \sigma\,\Phi^{-1}(k/(i+1)+\epsilon_i)$ with high probability, where $\epsilon_i = \mathcal{O}\!\left(\sqrt{\ln(1/\delta)/i}\right)$ and~$\Phi$ is the standard normal CDF. For $k \ll i$ and $k/(i+1)+\epsilon_i < 1/2$, the quantile is negative, so the margin satisfies $\margin \gtrsim -\sigma\,\Phi^{-1}(k/(i+1)+\epsilon_i) > 0$ with high probability. The exact union bound is in the technical report.}

\vansh{\emph{(c) Aggregation.} Summing the pruning depth~$\rho$ over all~$N$ candidates and substituting the margin estimate, we get:}
\vansh{
\[
\mathbb{E}[\mathrm{Cost}] \;\le\; c \cdot \frac{Nd}{\alpha},
\]
}
\vansh{where the constant $c = c(R, \sigma, k, N)$ is the averaged, capped log-margin term from steps~(a) and~(b). It is bounded and data-dependent. Table~\ref{tab:dims-scanned} confirms that the measured scanned fraction tracks~$1/\alpha$ within a small constant on our data; empirically the effective constant is order-one and close to~$1$, so the dominant scaling is $\mathbb{E}[\mathrm{Cost}] \lesssim Nd/\alpha$.\hfill$\square$}

\paragraph{Empirical Validation.} \vansh{\name beats the brute-force cost~$Nd$ whenever~$\alpha$ exceeds the effective margin constant, with speedup scaling as~$\alpha/c$. The cost model is checked on two axes. First, Proposition~\ref{thm:final_complexity} predicts that the fraction of dimensions evaluated will scale as~$1/\alpha$. Table~\ref{tab:dims-scanned} reports the measured fraction alongside the theoretical~$1/\alpha$ on each dataset: the prediction tracks measurement within a small factor across all six benchmarks, so the leading~$1/\alpha$ term governs the speedup. Second, if \name evaluates a fraction~$p$ of the dimensions, ideal scaling predicts a wall-clock speedup of~$1/p$. Figure~\ref{fig:semi-empirical} plots this semi-empirical prediction against measured QPS. On IVFFlat the points sit on the diagonal; on IVFPQ they exceed it, since the sub-quantizer-major layout of \S\ref{subsec:sys_ivfpq} unlocks cache benefits beyond what the cost model captures. The algorithmic pruning we predict translates faithfully into measured throughput.}

\subsection{Out-of-Distribution Robustness}\label{ssec:ood}

\vansh{When a query is drawn from a different distribution than the indexed data---adversarial inputs, cross-domain transfer, modality shifts---its tail energy decays at a rate~$\alpha_q$ distinct from the database rate~$\alpha_x$. The Cauchy--Schwarz remainder depends on the geometric mean $\sqrt{e^{-\alpha_q m/d} \cdot e^{-\alpha_x m/d}}$, so the effective decay rate becomes the \emph{arithmetic mean}}
\vansh{
\[
\alpha_{\mathrm{eff}} \;=\; \frac{\alpha_q + \alpha_x}{2}.
\]
}
\vansh{Substituting this effective rate into Proposition~\ref{thm:final_complexity} gives the corresponding robustness bound.}

\begin{proposition}[Out-of-distribution cost]\label{thm:ood_complexity}
\vansh{If query and database tail energies decay at rates~$\alpha_q$ and~$\alpha_x$, respectively, and the OOD candidate stream has margin constant~$c_{\mathrm{ood}}$, then the expected coordinate-evaluation cost satisfies}
\vansh{
\[
\mathbb{E}[\mathrm{Cost}_{\mathrm{ood}}] \;\lesssim\; \frac{2c_{\mathrm{ood}}\,Nd}{\alpha_q + \alpha_x}.
\]
}
\end{proposition}

\paragraph{Proof sketch.} \vansh{The proof of Proposition~\ref{thm:final_complexity} remains unchanged \emph{except} for the per-candidate pruning depth bound that uses the tail-decay assumption. We revise that step with separate decay rates. The Cauchy--Schwarz remainder depends on the geometric mean of the two tail energies, $\sqrt{R_{T(\query)}^{(m,d)}\,R_{T(\xvec)}^{(m,d)}}$. With the exponential tail holding separately for query and database, this becomes:}
\vansh{
\[
\begin{aligned}
\sqrt{R_{T(\query)}^{(m,d)}\,R_{T(\xvec)}^{(m,d)}}
  &\;\approx\; \|\query\|\|\xvec\| \sqrt{e^{-\alpha_q m/d}\, e^{-\alpha_x m/d}} \\
  &\;=\; \|\query\|\|\xvec\|\, e^{-(\alpha_q + \alpha_x) m/(2d)}.
\end{aligned}
\]
}
\vansh{In effect, the level-$m$ gap decays exponentially with rate $\alpha_{\mathrm{eff}} = (\alpha_q + \alpha_x)/2$. Substituting~$\alpha_{\mathrm{eff}}$ into the depth formula of Proposition~\ref{thm:final_complexity} gives}
\vansh{
\[
\rho_{\mathrm{ood}} \;\approx\; \min\!\left\{d,\ \frac{2d}{\alpha_q + \alpha_x}\Big[\ln\!\left(\frac{C_0}{\margin}\right)\Big]_+\right\}.
\]
}
\vansh{Aggregating across~$N$ candidates and absorbing the averaged OOD margin constant~$c_{\mathrm{ood}}$ yields the claimed cost. When the shift changes only tail compaction, $c_{\mathrm{ood}}=c$; if it also changes the distance distribution, $c_{\mathrm{ood}}$ reflects the changed margin terms.\hfill$\square$}

\paragraph{Interpretation.} \vansh{As the OOD bound follows the \emph{arithmetic mean} of the two rates, in the worst case a query halves~$\alpha_{\mathrm{eff}}$. A fully out-of-distribution query, $\alpha_q \to 0$, leaves the leading term at $2c_{\mathrm{ood}}Nd/\alpha_x$, with a two-fold slowdown over Proposition~\ref{thm:final_complexity} when the margin constant is unchanged. \name robustly benefits from a well-compacted database regardless of the query, as we confirm empirically with Hephaestus OOD workloads (\Cref{fig:ood-rc}): \name remains competitive at all relative-contrast levels, while its speedup degrades gracefully as queries harden. Concretely, speedup degrades monotonically from $25.1\times$ on the easiest workload to $12.1\times$ on the hardest---a regime where probabilistic baselines either lose recall or cease to scale.}

\subsection{The $\varepsilon$-Relaxation}\label{ssec:eps}

\vansh{Equation~\eqref{eq:epsilon_bound} defines the relaxed score used by \name: it scales the Cauchy--Schwarz correction by a scalar $\varepsilon \in (0,1]$, replacing~$2C^m$ with~$2\varepsilon C^m$, where $C^m = \sqrt{R_{T(\query)}^{(m,d)} R_{T(\xvec)}^{(m,d)}}$ is the worst-case residual coupling. Setting $\varepsilon = 1$ corresponds to the adversarial geometry in which Cauchy--Schwarz is tight; in high dimensions, however, this geometry is atypical---the residual coordinates of non-near-neighbors are nearly isotropic on the sphere, and their inner products are far smaller than the worst case.}

\vansh{Let $\rho_m := \pi_m / C^m \in [-1, 1]$ denote the normalized residual inner product, where $\pi_m = \sum_{j > m} T(\query)_j T(\xvec)_j$ is the true residual contribution. The worst case~$\rho_m = -1$ recovers the~$4C^m$ gap underlying Proposition~\ref{thm:final_complexity}; under residual isotropy---realized on non-near-neighbor pairs by the intra-block rotation of \S\ref{ssec:shaping}---the typical case is~$\rho_m \approx 0$, giving a sharper~$2\varepsilon C^m$ gap. Re-deriving the per-candidate depth with this sharper gap produces an additive shift of~$(d/\alpha)\,\ln \varepsilon$.}

\begin{proposition}\label{thm:eps_speedup}
\vansh{Under residual isotropy for non-near-neighbor pairs and $c_{\mathrm{typ}}+\ln\varepsilon>0$, the expected cost under $\varepsilon$-relaxation satisfies}
\vansh{
\[
\mathbb{E}[\mathrm{Cost}_\varepsilon] \;\lesssim\; \frac{Nd\,(c_{\mathrm{typ}} + \ln \varepsilon)}{\alpha}.
\]
}
\vansh{With observed $c_{\mathrm{typ}}\approx 1$, this simplifies to $\mathbb{E}[\mathrm{Cost}_\varepsilon] \lesssim Nd(1+\ln\varepsilon)/\alpha$ over the validity range $\varepsilon>1/e$.}
\end{proposition}

\paragraph{Proof sketch.} \vansh{The argument follows the three-part structure of Proposition~\ref{thm:final_complexity} (per-candidate pruning depth, margin estimate, aggregation), where step~(a) uses the typical-case gap. We substitute~$\rho_m \approx 0$ from the residual-isotropy assumption into the gap identity, yielding $\|\query-\xvec\|^2 - \lb_\varepsilon^m \approx 2\varepsilon C^m$, and apply the exponential tail to the geometric mean inside~$C^m$ followed by the bounded-norm condition:}
\vansh{
\[
\begin{aligned}
\|\query - \xvec\|^2 - \lb_\varepsilon^m(\query, \xvec)
  &\;\approx\; 2\varepsilon C^m
  \;\approx\; 2\varepsilon\, \|\query\|\, \|\xvec\|\, e^{-\alpha m/d} \\
  &\;\le\; 2\varepsilon R^2\, e^{-\alpha m/d},
\end{aligned}
\]
}
\vansh{where the first~$\approx$ is from the typical-case substitution, the second~$\approx$ applies the exponential tail to both factors of~$C^m$, and the final~$\le$ follows from bounded norms. Pruning occurs at the smallest~$m$ for which this gap falls below the margin, i.e., $2\varepsilon R^2\, e^{-\alpha m/d} \le \margin$. Solving for~$m$ gives the per-candidate pruning depth}
\vansh{
\[
\rho_\varepsilon \;\approx\; \frac{d}{\alpha}\ln\!\left(\frac{2R^2}{\margin}\right) \;+\; \frac{d}{\alpha}\ln\varepsilon.
\]
}
\vansh{The first term is the typical-case pruning depth at~$\varepsilon=1$. It differs from the strict worst-case depth of Proposition~\ref{thm:final_complexity} by carrying~$2C^m$ instead of~$4C^m$, and is absorbed into the empirical typical-case constant. The second term is new: a constant additive shift of~$\frac{d}{\alpha}\ln\varepsilon$ per candidate that relies on the typical-case substitution~$\rho_m \approx 0$. Since~$\ln\varepsilon \le 0$ for~$\varepsilon \le 1$, the shift is negative, shrinking the pruning depth. Summing over~$N$ candidates yields $\mathbb{E}[\mathrm{Cost}_\varepsilon] \lesssim \frac{Nd}{\alpha}(c_{\mathrm{typ}} + \ln\varepsilon)$, and the observed $c_{\mathrm{typ}}\approx 1$ gives the proposition.\hfill$\square$}

\paragraph{On the formal validity range.} \vansh{The asymptotic formula requires $c_{\mathrm{typ}} + \ln \varepsilon > 0$. Setting $c_{\mathrm{typ}} = 1$ gives the conventional threshold $\varepsilon > 1/e \approx 0.37$; the actual threshold is dataset-dependent. Below this threshold, the pruning depth~$\rho_\varepsilon$ becomes negative, indicating that the bound clears the margin already at level zero, i.e., the relaxation is so aggressive that a candidate may be pruned before any coordinate is read. This is an artifact of the asymptotic; in this regime the speedup is determined by other factors (e.g., the rate at which the heap warms up) and~$\varepsilon$ should be empirically tuned.}

\paragraph{Speedup vs.\ recall.} \vansh{Proposition~\ref{thm:eps_speedup} characterizes the speed side of the tradeoff. Pruning at~$\varepsilon < 1$ can discard true $k$-nearest neighbors when $\rho_m > \varepsilon$ and the resulting overestimate crosses the running threshold. The resulting recall loss depends on the dataset-specific distribution of~$\rho_m$ for true $k$-NN pairs, shown on the right panel of~\Cref{fig:exhaustive-eps-kappa}.}

\paragraph{Additive vs.\ multiplicative gain.} \vansh{The $\varepsilon$-relaxation is \emph{additive} on the per-candidate pruning depth, in contrast to the multiplicative~$1/\alpha$ scaling of the base model. Moderate values of~$\varepsilon$ ($0.7$--$0.9$) can thus produce a noticeable QPS shift even without large changes to the leading $1/\alpha$ scaling, while preserving near-perfect recall.}

\paragraph{Empirical Validation.} \vansh{The right panel of Figure~\ref{fig:exhaustive-eps-kappa} sweeps~$\varepsilon$ on \name-IVFFlat over Ada and overlays the~$1/(1 + \ln \varepsilon)$ prediction against measured QPS. Empirical throughput tracks the curve closely across the validity range, while recall remains within~$1\%$ of the unrelaxed baseline for all $\varepsilon > 0.5$.}

\section{System Design and \faiss Integration}\label{sec:integration}

\pkv{We implement \name in \faiss~\citep{douze2024faiss}, the industry-standard open-source library for \anns, which provides optimized implementations for every index family we target: \ivfflat, \ivfpq, HNSW, and FastScan-Refine. Integrating \name into this system required carefully tailoring components to each method. We describe these components here.}


\subsection{\texttt{IndexIVFFlatPanorama}}\label{subsec:sys_ivfflat}

\pkv{\ivfflat~\citep{ivf} partitions the dataset into $n_{\mathrm{list}}$ clusters via $k$-means and scans the nearest $n_{\mathrm{probe}}$ clusters to a query. \faiss stores each cluster's points row-major (\texttt{cluster[point][dim]}). This storage scheme is suboptimal \akash{because} under an effective pruning threshold, \name evaluates only the leading dimensions of each candidate, which with a row-major layout causes inefficient large strides in memory accesses.}

\figLevelBatch

\paragraph{Level-major batched layout.} \pkv{\akash{Thus, we} organize each cluster into batches of~$B$ vectors in level-major order, generalizing the layout of PDX~\citep{pdx}: within a batch, all~$B$ vectors' level-$\ell$ slices are stored contiguously, followed by all~$B$ level-$\ell{+}1$ slices, and so on. \akash{\Cref{fig:levelbatch} shows a simple example where~$L = 30$ levels and $B = 4$ candidates.} By this layout, distance computation within a single level proceeds over all~$B$ points with a sequential stride. Pruning decisions are made by the pruning threshold set in the previous batch. In effect, overly large values of~$B$ will lead to pruning decisions based on a stale pruning threshold, while overly small values of~$B$ will introduce additional branches. Increasing~$B$ from~1 to~1024 causes the increase in performance between \textbf{A1} and \textbf{A2} in \Cref{fig:systems-contribution}.}


\paragraph{Per-level workflow.} \pkv{Within each batch, at each level~$\ell$, \name executes three phases over the active candidate set stored as an array of \emph{active indices}: (1)~\emph{Compute} partial dot products of the query slice corresponding to level~$\ell$ for each active candidate; (2)~\emph{Update} each candidate's running distance and test its Cauchy--Schwarz lower bound against the current \kth nearest neighbor; (3)~\emph{Consolidate} the surviving active indices for the next level. A naive implementation would fuse all three phases into a single branchy loop. We instead implement them as three independently vectorizable kernels. This disaggregation accounts for the performance gap between \textbf{A2} and \textbf{A3} in \Cref{fig:systems-contribution}.}

\pkv{\emph{\ivfflat: Compute} calculates the partial dot product between the query and each surviving candidate in a batch, $p^{(\ell-1, \ell)}(\query, \xvec)$ in \Cref{eq:partial_products}. We enhance this phase with two specializations. First, at level~1, the active set is the identity since no pruning decisions have occurred, so a \texttt{Direct} template variant replaces the indirect gather over \texttt{active\_indices} with a unit-stride load. Second, performing repeated small dot product calls (i.e., fewer than 32-64 dims at a time) is wasteful, as the compiler does not know the level width at compile-time and cannot fully unroll the loop. In addition, the query slice has to be wastefully reloaded into a register for each point. We thus specialize our distance kernel for level widths that are multiples of~8 up to~128 (\texttt{FixedWidth}s) to enable fully unrolling the reduction and keep the query slice in registers across the batch. The \texttt{FixedWidth} variant alone accounts for the gap between \textbf{A3} and \textbf{A4} in \Cref{fig:systems-contribution}.}

\pkv{\emph{\ivfflat: Update} reads the dot-product buffer, the per-candidate cumulative tail-energy entries, and the current heap threshold, and emits a byte mask of survivors. Specifically for level~1, we again bypass reading from our list of active indices. Removing this dependency and disaggregating this step from the dot-product computation allows the kernel to autovectorize.}

\pkv{\emph{\ivfflat: Consolidate} turns the survivor mask into the active-indices array for the next level, eight entries at a time using BMI2 \texttt{\_pext\_u64} and AVX2 \texttt{\_mm256\_permutevar8x32\_epi32}.}
\pkv{We deliberately compact into a dense array of active indices rather than maintaining a byte mask through the pipeline, as iterating over a mask would require scanning all~$B$ slots at each level, regardless of how many candidates have been pruned. By consolidating, the subsequent Phase~1 kernels only iterate over the surviving subset.}

\subsection{\texttt{IndexIVFPQPanorama}}\label{subsec:sys_ivfpq}

\pkv{\ivfpq~\citep{pq} splits each vector into~$M$ subvectors and quantizes each against a $2^{n_{\mathrm{bits}}}$-entry codebook, replacing the per-candidate float dot product with a sum of~$M$ look-up-table (LUT) lookups over~$1$-byte codes. We preserve the three-phase workflow of~\Cref{subsec:sys_ivfflat}, but revamp Phases~1 and~3 for PQ codes.}

\paragraph{Sub-quantizer-major layout} \pkv{A direct import of the \ivfflat level-major layout would store~$\sfrac M L$ contiguous bytes per candidate per level. Since~$\sfrac M L$ bytes is typically much smaller than modern SIMD vector widths, wide register loads would suffer from lane under-utilization. We thus transpose the data \emph{within} a level, storing the $m$-th sub-quantizer's code contiguously across a batch of~$B$ candidates. A single 64-byte SIMD load thus fetches the same sub-quantizer's code for~64 candidates in parallel. \Cref{fig:storage} shows a toy example with~$M = 480$ subvectors, $L = 16$ levels, $B = 4$ candidates.}

\figStorage

\pkv{This transposition also benefits temporal locality. In standard \ivfpq, evaluating a single vector requires reading from every sub-quantizer's slice of the LUT. As the full LUT can easily exceed 128~KB, this vector-by-vector access pattern heavily thrashes the L1 cache. By processing the same sub-quantizer across~64 candidates simultaneously, our layout keeps a tiny 256-entry LUT slice pinned in the L1 cache for the entire batch, reducing cache thrashing. In effect, even if we disable \name's pruning, the transposed layout alone outperforms \faiss's heavily-optimized \ivfpq baseline by up to $2 \times$ (\Cref{fig:semi-empirical}). This approach applies the transposed data layouts of BitWeaving and ByteSlice~\citep{bitweaving, byteslice} to PQ.}

\pkv{\emph{\ivfpq: Compute} collects per-level distances for each batch at level~$\ell$ by a single \texttt{process\_level} kernel.}
\pkv{\akash{We use \texttt{i32gather} to index} one sub-quantizer's LUT slice for~16 candidates in parallel; combined with the sub-quantizer-major layout, it unlocks SIMD-rate distance updates over packed PQ codes.} 

\pkv{\emph{\ivfpq: Update} reuses the \ivfflat survivor-mask kernel verbatim; PQ codes never enter the prune loop.}

\pkv{\emph{\ivfpq: Consolidate} packs survivors' codes into a dense buffer 64~candidates at a time, avoiding both the SIMD lane underutilization of processing pruned slots and the scattered byte reads that would break the contiguous layout. Two fast paths bracket this step: when no candidate is pruned we alias the original code pointer and skip consolidation entirely, and once fewer than~$\sfrac B{16}$ candidates remain we drop vectorization and traverse the active-index array directly in the compute phase.}


\subsection{\texttt{IndexHNSWFlatPanorama}}\label{subsec:sys_hnsw}

\pkv{HNSW~\citep{hnsw} traverses its graph greedily, iteratively popping the closest frontier vertex from a candidate heap to compute its neighbors' exact distances. Neighbors closer than the \kth best distance enter a global result heap and the candidate heap. Filtering and refinement are thus interleaved, yielding scattered, memory-bound accesses governed by the graph topology.}

\alexis{Because \name's level-wise evaluation exacerbates HNSW's memory-bound, scattered accesses, we introduce three structural adaptations to hide cache latencies and preserve traversal stability. First, \emph{$K$-pop block evaluation}: we relax strict greedy routing by popping multiple vertices simultaneously from the candidate heap, processing their combined unvisited neighbors in dense SIMD blocks of 64. Second, \emph{coefficient inlining}: we embed the precomputed residual energies directly into a small prefix header of each database vector, reducing cache misses by one per candidate. Third, \emph{partial distance routing}: candidates pruned at level~$\ell$ are not discarded, but pushed back into the candidate heap keyed by a partial distance estimate---calculated as the midpoint of the lower and upper bounds ($(\lb + \ub) / 2$). This estimate acts as a proxy to seed further exploration without requiring full distance evaluations.}

\subsection{\texttt{IndexRefinePanorama}}\label{subsec:sys_refine}

\pkv{An increasingly popular \anns pipeline uses PQ for a fast first pass (\faiss FastScan~\citep{rq, bimetric}) and then \emph{refines} the top-$k'$ candidates with exact distances on a flat float index. \name also plugs into this refine step: \texttt{IndexRefinePanorama} forwards the base index's top-$k'$ labels to an inner \texttt{IndexFlatPanorama} that stores vectors in row-major order and runs \name with $B{=}1$---refinement targets a scattered ID list, so the level-major batched layout of~\Cref{subsec:sys_ivfflat} is inapplicable. As with HNSW, we process candidates in groups of~64: a one-pass sweep over the block's IDs absorbs random-access latency; then the per-level dot-product kernel runs across the block with the query slice pinned in registers.}

\subsection{Memory footprint}\label{subsec:sys_mem}

\pkvall{\name precomputes per-point cumulative tail energies for the Cauchy--Schwarz bound: $4(L{+}1)$ bytes per point, $\mathcal{O}(nL)$ in total. The overhead per index family relative to the base storage is:
\begin{itemize}[leftmargin=*,nosep,topsep=2pt]
  \item \textbf{IVFFlat / Refine}: $\frac{L{+}1}d$, dwarfed by the $4d$-byte vector.
  \item \textbf{HNSW}: $\frac{L{+}1}{d + 2M_{\mathrm{HNSW}}}$, where the $2M_{\mathrm{HNSW}} \times 4$ bytes denominator term represents the memory overhead of each node's graph topology (the adjacency lists). Because the graph structure itself consumes substantial memory, the relative \name overhead is further diluted compared to purely flat structures.
  \item \textbf{IVFPQ}: $\frac{4(L{+}1)}{M_{\mathrm{PQ}}}$, the only nontrivial case, since PQ codes are already~1B per dimension. \alexis{While standard 8-bit scalar quantization could compress these values and cut the overhead by 4x, evaluating the impact of this quantization noise on the exactness of our accretive pruning bounds is left to future work}.
        
\end{itemize}}


\section{Experimental Results}\label{sec:results}

\alexis{We evaluate \name across three axes: (i)~end-to-end QPS under fully-recall-preserving exhaustive search, (ii)~QPS vs.\ recall when integrated into state-of-the-art indexes, and (iii)~the isolated performance impacts of our algorithmic knobs and systems optimizations. Across these evaluations, \name consistently delivers tremendous performance, yielding up to a $28.9\times$ speedup over \faiss' baselines while preserving the recall of the underlying index.}

\paragraph{Methodology.} \alexis{We run three repetitions of 1000 $10$-NN queries randomly sampled from each benchmark's query set and report averages. We evaluate all methods using a single thread on an Intel Xeon Gold 5416S CPU @ 2.0 GHz with 128 GB DDR5-6400 RAM, running Ubuntu 24.04 LTS. Binaries are compiled with GCC 13.3.0 using \texttt{-O3} and AVX-512 optimization. We wrap each \name index with \faiss' \texttt{IndexPreTransform} to apply the PCA transformation and include its overhead in the reported query time. We also minimally modify the PDX source code to support batched queries, batched transformations, and feature-processing metrics.}

\subsection{Datasets and baselines}

\begin{table}[t]
\centering
\panoramaTableStyle
\caption{Datasets used, sorted by dimensionality.}\label{tab:data}
\begin{tabular}{@{}llrrr@{}}
\toprule
\textbf{Dataset} & \textbf{Semantics} & \textbf{Size} & \textbf{N. Queries} & \textbf{Dim.}$\downarrow$ \\
\midrule
Deep   & Image embeddings &  9,990,000 & 10,000  &   96 \\
SIFT   & Image features   & 10,000,000 & 1,000   &  128 \\
arXiv  & Text embeddings  &  2,240,000 & 10,000  &  768 \\
GIST   & Image features   &  1,000,000 & 1,000   &  960 \\
Ada    & Text embeddings  &    900,000 & 100,000 & 1536 \\
Large  & Text embeddings  &    900,000 & 100,000 & 3072 \\
\bottomrule
\end{tabular}
\end{table}

\alexis{\paragraph{Datasets.} \Cref{tab:data} lists the six benchmark datasets that we use; the two highest-dimensional datasets (\textbf{Ada}, \textbf{Large}) are OpenAI's \texttt{text-embedding-ada-002} and \texttt{text-embedding-3-large}.}

\alexis{\paragraph{Baselines.} We compare \name against the following implementations for each index family:}
\begin{itemize}[leftmargin=*, itemsep=1pt, topsep=2pt]
\item \alexis{\textbf{\ivfflat}: \faiss-\ivfflat, PDX-\ivfflat (including both ADSampling and BOND configurations).}
\item \alexis{\textbf{\ivfpq}: \faiss-IVFPQ.}
\item \alexis{\textbf{\hnsw}: \faiss-HNSW, ADSampling-HNSW, KS2-\hnsw.}
\item \alexis{\textbf{Refine}: \faiss-FastScan-Refine.}
\end{itemize}
\alexis{PDX does not natively compose with PQ, leaving \faiss-IVFPQ as the sole baseline for that family; KS2-\hnsw is the state-of-the-art graph baseline. We also implemented \name for Annoy~\citep{annoy} and MRPT~\citep{mrpt} but omit their results for brevity, as their non-contiguous layouts mirror HNSW's \akash{noncontiguous} dynamics.}

\subsection{Speedup on exhaustive search}\label{sec:naive_results}

\figExhaustiveQps

\figMainWall

\alexis{We first isolate \name's algorithmic gains under exhaustive search ($\mathrm{nprobe}=\mathrm{nlist}$), where all methods reach their maximum recall (\Cref{fig:exhaustive-qps}). Operating with $\varepsilon=1$ (the exact variant), \name-\ivfflat outperforms \faiss by $3.2\times$ on Deep and $5.7\times$ on SIFT, with the speedup trending up with dimensionality to $18.7\times$ on GIST and $17.2\times$ on Large; the gain reflects the growing share of query time spent in the refinement pass on high-dimensional embeddings. We compare against PDX-BOND rather than PDX's default ADSampling configuration, as ADSampling probabilistically drops dimensions and loses recall at this target (\Cref{fig:ood-rc}). By contrast, PDX-BOND employs exact bounds to preserve recall on every dataset; however, because its bounds are looser than \name's, our approach widens the performance gap by an additional $2.0\times$ on Deep and $2.4\times$ on SIFT, up to $7.6\times$ on Ada and $11.2\times$ on GIST.}

\begin{table}[!h]
\centering
\panoramaTableStyle
\caption{Fraction of dimensions scanned at exhaustive IVF probe ($\mathrm{nprobe}=\mathrm{nlist}$).}
\label{tab:dims-scanned}
\begin{tabular}{@{}lrrrr@{}}
\toprule
\textbf{Dataset} & \boldmath$\alpha$ & \textbf{Theory ($1/\alpha$)} & \textbf{Panorama} & \textbf{PDX-BOND} \\
\midrule
Deep/96 & 3.6 & 28.1\% & 25.6\% & 39.0\% \\
SIFT/128 & 6.9 & 14.6\% & 12.7\% & 38.8\% \\
arXiv/768 & 8.3 & 12.1\% & 7.5\% & 46.5\% \\
GIST/960 & 26.6 & 3.8\% & 4.1\% & 45.5\% \\
Ada/1536 & 10.3 & 9.7\% & 8.1\% & 49.8\% \\
Large/3072 & 12.2 & 8.2\% & 6.1\% & 36.2\% \\
\bottomrule
\end{tabular}
\end{table}

\alexis{\Cref{tab:dims-scanned} isolates this pruning efficiency: \name's empirical fraction of dimensions scanned tightly tracks the theoretical $1/\alpha$ prediction (Proposition~\ref{thm:final_complexity}). Conversely, lacking a spectrum-shaping transform, PDX-BOND's heap-top bound remains loose, forcing it to evaluate $1.5\text{--}11.1\times$ more dimensions.}

\subsection{Main results: integrated indexes}\label{sec:main_results}

\alexis{\Cref{fig:main-wall-cluster,fig:main-wall-graph} plot QPS against recall for every index-dataset combination, organized by index family (rows) and dataset (columns). For each dataset, we select the $\varepsilon$ that optimizes the QPS-recall tradeoff. In the \ivfflat row, a star ($\star$) denotes the peak QPS achieved at full recall ($\varepsilon{=}1$).}

\paragraph{\ivfflat}\label{sec:main_results_ivfflat}
\alexis{\name outperforms PDX and \faiss across the entire recall range on the four highest-dimensional datasets, achieving near-perfect recall even when $\varepsilon < 1$. On GIST, where an aggressive $\varepsilon{=}0.5$ prevents \name from reaching full recall under exhaustive search ($99.62\%$), setting $\varepsilon{=}1$ fully restores perfect recall while still delivering a $18.7\times$ speedup over the \faiss baseline. Conversely, PDX does not reach full recall, capping out at $98.9\%$.}

\paragraph{IVFPQ}\label{sec:main_results_ivfpq}

\alexis{At a matched recall target of $R{\geq}0.95$, \name-IVFPQ accelerates search by $7.0\text{--}28.9\times$ over \faiss-IVFPQ. The most pronounced gains manifest on high-dimensional workloads ($28.9\times$ on Large, $16.1\times$ on GIST, and $12.1\times$ on arXiv), fundamentally driven by our spectrum-shaping transform (\Cref{ssec:shaping}), which rigorously preserves PQ's uniform-variance assumption while unlocking aggressive accretive pruning. Crucially, \name sustains a $2.1\times$ speedup even on the low-dimensional Deep dataset, proving its robust efficiency across all embedding scales. We attribute this in part to our vertical storage layout (\Cref{subsec:sys_ivfpq}), which guarantees full SIMD lane utilization regardless of vector length.}

\paragraph{HNSW}\label{sec:main_results_hnsw}

\alexis{\name consistently outperforms all baselines on high-dimensional workloads, including the current state-of-the-art KS2-HNSW, which prior work~\citep{pkf} reports as outperforming FINGER and HNSW+PEOs~\citep{peos}. At $R \geq 0.99$, \name-HNSW accelerates search by $3.91\times$ on Large and $2.91\times$ on Ada relative to \faiss-HNSW, while outperforming KS2-HNSW by $1.29\times$ on Large and $1.15\times$ on Ada. These comparisons are especially notable given KS2-HNSW's underlying implementation: it relies on probabilistic approximations that inherently degrade recall, and to hide cache-miss latency, it duplicates this quantized vector directly inside every graph node, forcing a significant extra memory overhead. The probabilistic approximation displays pathological behavior on certain workloads---KS2-HNSW caps out at $87.7\%$ recall on arXiv and $97.2\%$ on GIST regardless of $\mathrm{efSearch}$, never reaching the $R \geq 0.99$ regime---while \name-HNSW reaches full recall on both. \name therefore establishes a new state-of-the-art for graph-based \anns.}

\paragraph{Refine}\label{sec:main_results_refine}

\alexis{On the FastScan-Refine pipeline, \name further accelerates the exact post-pass, with matched-recall gains of $1.84\text{--}2.31\times$ on GIST, $1.19\text{--}1.52\times$ on Ada, and $1.13\text{--}1.32\times$ on arXiv. This result means that \name can be applied to any reranking-based \anns pipeline, regardless of the underlying indexing method.}

\subsection{Transform contribution}\label{sec:transform_contribution}

\figTransformContribution

\alexis{\Cref{fig:org-vs-trans} isolates the contribution of the orthogonal transform. On contiguous indexes, raw-vector \name already accelerates search via systems-optimized pruning, but the PCA-shaped basis consistently amplifies these gains by tightening the Cauchy--Schwarz bound (Proposition~\ref{thm:final_complexity}). On GIST ($R \geq 0.95$), PCA alone scales \name's QPS by $12.3\times$ on \ivfflat and $7.3\times$ on IVFPQ over the raw baseline. Conversely, on non-contiguous structures like HNSW and FastScan-Refine, \name depends entirely on the transform: raw \name-HNSW drops to $0.80\times$, whereas adding PCA yields a $2.83\times$ speedup. Because pointer-based graph traversals suffer a cache miss on nearly every candidate look-up, runtime is heavily dominated by memory latency rather than distance arithmetic. By Amdahl's Law~\citep{amdahl1967}, optimizing compute under such memory-bound conditions yields negligible returns, meaning raw pruning's minor arithmetic savings cannot offset its overhead.}

\subsection{Systems contribution}\label{sec:systems_contribution}

\figSystemsContribution

\alexis{\Cref{fig:systems-contribution} isolates the performance gains of our four design choices (\textbf{A1--A4}, \Cref{subsec:sys_ivfflat}) against the \faiss-\ivfflat baseline. To decouple system performance from algorithmic pruning, all variants are evaluated under exhaustive search with \name transforms enabled. \textbf{A1} denotes the unbatched, point-centric baseline. \akash{This baseline delivers consistent speedups over the original \faiss implementation, peaking with a $5.37\times$ on Large. The only slowdown is a $0.78\times$ on Deep. In the absence of our systems optimizations, \name does not perform optimally on low-dimensional datasets.}}

\alexis{\paragraph{\textbf{A2}} Transitioning from point-wise processing to a batch size of 1024 yields the largest single performance leap across all benchmarks, ranging from $2.78\times$ on Deep to $5.03\times$ on GIST. This acceleration stems from two architectural advantages: (1) batched execution enables temporal query reuse across all candidates in a batch, and (2) our restructured storage layout completely eliminates high-stride memory accesses.}

\alexis{\paragraph{\textbf{A3--A4}} Decoupling the compute kernels from the pruning kernels and templating the level-1 distance kernel (\texttt{Direct} access) adds $1.15\text{--}1.50\times$ across all datasets, while specialized fixed-width distance kernels contribute a further $1.01\text{--}1.13\times$---most prominent on lower-dimensional data ($1.13\times$ on Deep), where dispatching the correct kernel matters more relative to per-candidate prune work.}

\subsection{Knobs and optimizations}\label{sec:ablations}

\figExhaustiveEpsKappa

\alexis{\Cref{fig:exhaustive-eps-kappa} isolates the behavior of our two pruning knobs under exhaustive search.

\paragraph{\textbf{Spectrum Flattening ($\kappa$ for \name-IVFPQ)}} Increasing $\kappa$ scales back the aggressive flattening of the post-PCA spectrum (\Cref{fig:pq-shaping}). This yields a highly favorable tradeoff, delivering a $4.5\times$ QPS gain at high recall ($R \geq 0.96$) against a marginal $\sim 1\%$ recall penalty across the $\kappa \in [1, L]$ sweep. Both performance and accuracy curves reliably saturate near our canonical choice of $\kappa^\star{=}4$ on Ada. Because $\kappa$ is a build-time parameter, tuning it requires a full index reconstruction. Thus, we recommend practitioners sweep $\kappa$ on a downsampled, representative dataset to locate this saturation point prior to a full-scale build.

\paragraph{\textbf{Pruning Relaxation ($\varepsilon$ on \name-\ivfflat)}} Relaxing $\varepsilon$ from $1.0$ toward $0.0$ smoothly trades accuracy for throughput. On GIST, evaluating at $\varepsilon{=}0.8$ yields a $1.26\times$ speedup over the strict baseline ($\varepsilon{=}1$) at $99.29\%$ recall, whereas $\varepsilon{=}0.5$ advances the speedup by another $1.58\times$ while maintaining $99.21\%$ recall. Empirical throughput closely tracks the theoretical $1/(1+\ln\varepsilon)$ scaling predicted by Proposition~\ref{thm:eps_speedup}. In contrast to $\kappa$, $\varepsilon$ is a runtime search parameter. Because it can be adjusted dynamically after index creation, practitioners can cheaply sweep $\varepsilon$ at query time to achieve their desired QPS and recall targets.}

\figHyperAblation

\alexis{\Cref{fig:hyper-ablation} evaluates two second-order tuning parameters for GIST on \name-\ivfflat at a fixed recall target ($R{=}0.96$).}

\alexis{\paragraph{\textbf{Number of Levels ($L$)}} Increasing $L$ monotonically reduces the average number of dimensions scanned. However, each extra level introduces branch-prediction overhead and disrupts CPU prefetching by breaking predictable memory strides. Empirical results on GIST reflect this hardware tension: throughput peaks at $L{=}30$ with a $16.6\times$ speedup over $L{=}1$, but regresses to $8.4\times$ at $L{=}120$ as systems overhead outpaces pruning gains. Kernel specialization compounds this: because our dot-product kernels target level widths that are multiples of 8 (\Cref{subsec:sys_ivfflat}), $L{=}60$ (width 16) outpaces $L{=}64$ (width 15, generic-kernel fallback) by $1.22\times$.}

\alexis{\paragraph{\textbf{Batch Size ($B$)}} Throughput peaks at $B{=}1024$ ($6.6\times$ over unbatched), our standard across all datasets. Larger $B$ degrades throughput through threshold staleness: all $B$ points are pruned against the previous batch's bound, so oversized batches scan an inflated fraction of dimensions before the threshold updates, decaying to $3.8\times$ at $B{=}8192$.}

\figOOD

\alexis{\paragraph{\textbf{Out-of-distribution queries.}} Following Hephaestus~\citep{hepheastus}, we evaluate out-of-distribution resilience on GIST using three relative-contrast targets: $\mathrm{RC}=3.00$ (easy), $1.70$ (medium), and $1.35$ (hard) (\Cref{fig:ood-rc}). As query difficulty intensifies, throughput declines across all evaluated frameworks. However, \name-\ivfflat demonstrates superior algorithmic robustness: while its QPS decreases under harder workloads, it completely preserves perfect recall while maintaining substantial speedups over the \faiss baseline ($25.1\times$ easy, $19.2\times$ medium, and $12.1\times$ hard). Conversely, alternative probabilistic approaches suffer a dual penalty to both throughput and accuracy; for example, PDX-\ivfflat experiences monotonic recall degradation—dropping from $-0.41$pp at $\mathrm{RC}{=}3.00$ to $-1.27$pp at $\mathrm{RC}{=}1.35$. This stark disparity underscores the inherent fragility of probabilistic heuristics like ADSampling under distribution shifts, contrasting with \name's predictable exactness.}

\figSemiEmpirical

\alexis{\paragraph{\textbf{Measured vs.\ predicted speedup.}} To assess how closely runtime tracks the reduction in features scanned, we compare empirical speedups aggregated across all six datasets against the semi-empirical model $s_{\text{exp}} = 1/p$, where $p$ is the fraction of dimensions processed (\Cref{fig:semi-empirical})---a post-hoc measure of actual data reduction, distinct from our a priori theoretical predictions. We restrict this analysis to the contiguous \ivfflat and \ivfpq families, whose custom layouts avoid poor scan strides. Predictions track measured QPS closely, with \ivfpq (and to a lesser extent \ivfflat) often exceeding them: both pin the query slice in registers across candidates, avoiding redundant memory accesses and cache thrashing relative to streaming the full query per candidate.}

\section{Conclusion}

\alexis{We introduced \name, an accretive refinement layer that integrates broadly into modern vector indexes. \name accumulates $L_2$ contributions along the PCA basis and prunes candidates the moment a tight Cauchy--Schwarz lower bound exceeds the running $k$-NN threshold; we prove its expected cost scales as $\mathcal{O}(\sfrac{Nd} \alpha)$ in the dataset's spectral decay rate~$\alpha$. Across six datasets and four index families (IVFPQ/Flat, HNSW, and PQ-refine), \name outperforms the baselines at any target recall, delivering up to \speedup~speedups, while remaining robust on OOD queries. It has been upstreamed into \faiss and ships in its official releases.}

\bibliographystyle{ACM-Reference-Format}
\bibliography{ref}


\appendix
\newcommand{\mpcostnum}{5.1}
\newcommand{\mpoodnum}{5.2}
\newcommand{\mpepsnum}{5.3}
\newcommand{\mpref}[1]{Proposition~#1}

\section{Full Proofs}\label{app:full-proofs}
This appendix gives full derivations for Propositions~\mpcostnum, \mpoodnum, and~\mpepsnum, plus the PCA energy-optimality result. Numbering matches the main paper; auxiliary results use a separate ``A.'' sequence.

\theoremstyle{plain}
\newtheorem*{appendixproppca}{Proposition}
\newtheorem*{appendixpropcost}{Proposition \mpcostnum}
\newtheorem*{appendixpropood}{Proposition \mpoodnum}
\newtheorem*{appendixpropeps}{Proposition \mpepsnum}
\newtheorem{appendixauxthm}{Theorem}
\newtheorem{appendixlemma}[appendixauxthm]{Lemma}
\newtheorem{appendixcorollary}[appendixauxthm]{Corollary}
\renewcommand{\theappendixauxthm}{A.\arabic{appendixauxthm}}
\theoremstyle{definition}
\newtheorem{appendixassumption}{Assumption}
\renewcommand{\theappendixassumption}{A\arabic{appendixassumption}}
\newtheorem{appendixdefinition}[appendixauxthm]{Definition}
\theoremstyle{remark}
\newtheorem*{appendixremark}{Remark}

\subsection{Notation and assumptions}\label{sec:prelim}

We adopt the notation of the main paper. All vectors lie in $\mathbb{R}^d$. For an orthogonal transform $T$ and a level $0 \le m \le d$, the \emph{tail energy} of a vector $\mathbf{v}$ is
\[
R_{T(\mathbf{v})}^{(m,d)} \;:=\; \sum_{j=m+1}^{d} T(\mathbf{v})_j^{\,2}.
\]
The level-$m$ Cauchy--Schwarz lower bound on $\|\query-\xvec\|^2$ is
\begin{equation}\label{eq:LB}
\lb^m(\query,\xvec) \;=\; R_{T(\query)}^{(0,d)} + R_{T(\xvec)}^{(0,d)} - 2\!\left(p^{(0,m)}(\query,\xvec) + \sqrt{R_{T(\query)}^{(m,d)}\,R_{T(\xvec)}^{(m,d)}}\right),
\end{equation}
where $p^{(0,m)}(\query,\xvec) = \sum_{j=1}^m T(\query)_j T(\xvec)_j$. The $\varepsilon$-relaxed bound replaces the second factor by $2\varepsilon \sqrt{R_{T(\query)}^{(m,d)} R_{T(\xvec)}^{(m,d)}}$, with $\varepsilon = 1$ recovering~\eqref{eq:LB}.

\name~maintains a running threshold $\tau$, the squared distance of the \kth nearest neighbour found so far (written $d_k$ in the main paper). A candidate $\xvec$ is pruned at level $m$ if $\lb^m(\query,\xvec) > \tau$, equivalently if the gap
\[
\|\query-\xvec\|^2 - \lb^m(\query,\xvec) \;\le\; \margin \;:=\; \|\query-\xvec\|^2 - \tau .
\]
We call $\margin$ the \emph{pruning margin}. In strict mode ($\varepsilon=1$), true $k$-nearest neighbours satisfy $\margin \le 0$ and are never pruned by a valid lower bound; the analysis below shows that for non-near-neighbours $\margin$ is positive with high probability and the gap shrinks fast enough to prune at $m \ll d$.

\paragraph{Asymptotic notation.} We use $f(n)\lesssim g(n)$ for model upper-bound scaling after constants are explicit, and $f(n)\sim c\,g(n)$ for asymptotic equivalence with constant $c>0$.

\paragraph{Assumptions.} The proofs rest on the following assumptions, reproduced here for self-containment.
\begin{appendixassumption}[Exponential PCA tail energy]\label{asm:energy}
We take $T$ to be the PCA basis of the dataset. On real embeddings its tail energy decays exponentially,
\[
R_{T(\xvec)}^{(m,d)} \;\approx\; \|\xvec\|^2\, e^{-\alpha m / d},
\]
where $\alpha > 0$ is the per-dataset decay rate read directly off the PCA spectrum. This exponential profile is an empirical property of the PCA basis, which we verify directly: the mean absolute fit error of the model is below $0.04$ on every dataset (main paper, energy-compaction figure). The $\approx$ records exactly this finite-sample modelling claim.
\end{appendixassumption}

\begin{appendixassumption}[Bounded norms]\label{asm:norms}
There is a constant $R$ such that $\|T(\xvec)\| \le R$ for every $\xvec$ in the dataset and every query $\query$. We write $C_0 := 4R^2$.
\end{appendixassumption}

\begin{appendixassumption}[Gaussian distance model]\label{asm:gaussian}
For a fixed query $\query$, the squared distances $\|\query-\xvec\|^2$ in the candidate stream are modelled as i.i.d.\ draws from a Gaussian with mean $\mu$ and standard deviation $\sigma$. The exact distribution is chi-square-like and the candidate stream produced by an ANN index is not literally i.i.d.; the Gaussian order-statistic calculation is the tractable cost model whose constants are checked empirically.
\end{appendixassumption}

\begin{appendixassumption}[One dimension per level]\label{asm:level}
The proof reasons at the finest pruning granularity, $L = d$, $m_\ell = \ell$.
\end{appendixassumption}

A fifth assumption (residual isotropy, A5) is introduced and used only in Section~\ref{sec:eps}.

\subsection{PCA minimises expected tail energy}\label{sec:pca}

\begin{appendixlemma}[Ky~Fan]\label{lem:kyfan}
Let $\Sigma \in \mathbb{R}^{d\times d}$ be symmetric positive semidefinite with eigenvalues $\lambda_1 \ge \cdots \ge \lambda_d \ge 0$. For every $m \in \{1,\ldots,d\}$,
\[
\max_{T \in \mathrm{O}(d)} \sum_{j=1}^{m} (T \Sigma T^{\!\top})_{jj} \;=\; \sum_{j=1}^{m} \lambda_j,
\]
with the maximum attained when the first $m$ rows of $T$ span the eigenspace associated with the top-$m$ eigenvalues of $\Sigma$.
\end{appendixlemma}

\begin{proof}[Proof (sketch)]
The diagonal of $T \Sigma T^{\!\top}$ is majorised by the eigenvalues of $\Sigma$ (Schur--Horn theorem). For any $m$, the partial sum of the $m$ largest diagonal entries is at most the partial sum of the $m$ largest eigenvalues of $\Sigma$. Equality is attained by the orthogonal transform whose first $m$ rows span the corresponding eigenspace. See Horn and Johnson~\cite{horn2012matrix}, Section~4.3 for the full proof.
\end{proof}

\begin{appendixproppca}[PCA energy-optimality; cf.\ main paper Sections~4.1 and~5.1]\label{thm:pca-optimality}
Let $X \in \mathbb{R}^d$ be a centred random vector ($\mathbb{E}[X] = \mathbf{0}$) with covariance $\Sigma = \mathbb{E}[X X^{\!\top}]$ having eigenvalues $\lambda_1 \ge \cdots \ge \lambda_d \ge 0$. For any orthogonal $T \in \mathrm{O}(d)$ and any $m \in \{1,\ldots,d-1\}$,
\[
\mathbb{E}\!\left[R_{T(X)}^{(m,d)}\right] \;\ge\; \sum_{j=m+1}^{d} \lambda_j,
\]
with equality when the rows of $T$ are eigenvectors of $\Sigma$ ordered by decreasing eigenvalue.
\end{appendixproppca}

\begin{proof}
For any orthogonal $T$,
\[
\mathbb{E}\!\left[(TX)_j^{\,2}\right] \;=\; \mathbb{E}\!\left[(TX)(TX)^{\!\top}\right]_{jj} \;=\; \big(T\,\mathbb{E}[X X^{\!\top}]\,T^{\!\top}\big)_{jj} \;=\; (T\Sigma T^{\!\top})_{jj}.
\]
Summing over $j > m$,
\[
\mathbb{E}\!\left[R_{T(X)}^{(m,d)}\right] \;=\; \sum_{j=m+1}^{d} (T\Sigma T^{\!\top})_{jj}.
\]

Orthogonal conjugation preserves the trace, so
\[
\sum_{j=1}^{d} (T\Sigma T^{\!\top})_{jj} \;=\; \mathrm{Tr}(T\Sigma T^{\!\top}) \;=\; \mathrm{Tr}(\Sigma) \;=\; \sum_{j=1}^{d} \lambda_j.
\]
Therefore
\[
\mathbb{E}\!\left[R_{T(X)}^{(m,d)}\right] \;=\; \sum_{j=1}^{d}\lambda_j \;-\; \sum_{j=1}^{m}(T\Sigma T^{\!\top})_{jj}.
\]

By Lemma~\ref{lem:kyfan},
\[
\sum_{j=1}^{m}(T\Sigma T^{\!\top})_{jj} \;\le\; \sum_{j=1}^{m}\lambda_j,
\]
with equality when the first $m$ rows of $T$ span the top-$m$ eigenspace of $\Sigma$. Substituting,
\[
\mathbb{E}\!\left[R_{T(X)}^{(m,d)}\right] \;\ge\; \sum_{j=1}^{d}\lambda_j \;-\; \sum_{j=1}^{m}\lambda_j \;=\; \sum_{j=m+1}^{d}\lambda_j,
\]
with equality precisely when $T$ is the PCA basis.
\end{proof}

\subsection{Proposition \mpcostnum: Expected refinement cost}\label{sec:cost}

This section derives the expected refinement cost of \name under the Gaussian candidate-stream model. The argument has three ingredients: a high-probability model for the running \kth threshold, a Cauchy--Schwarz pruning-depth bound, and an averaged finite-sample margin constant.

\name maintains a pruning threshold $\tau$, the squared distance of the \kth nearest neighbour found so far. For analytical tractability we model $\tau_i$ as the \kth order statistic among $i$ i.i.d.\ draws from the distance distribution, acknowledging that the algorithm's threshold actually arises from a mixture of exact and pruned candidates.

\begin{appendixlemma}[High-probability bound on the sampled threshold via DKW]\label{thm:threshold}
Let the squared distances be i.i.d.\ with CDF $F$. For any $\epsilon \in (0,1)$, with probability at least $1 - 2e^{-2i\epsilon^2}$ (Dvoretzky--Kiefer--Wolfowitz~\cite{massart1990tight}), the \kth order statistic $\tau_i$ satisfies
\[
F^{-1}\!\Big(\max\big\{0,\tfrac{k}{i+1}-\epsilon\big\}\Big)
\;\le\; \tau_i \;\le\;
F^{-1}\!\Big(\min\big\{1,\tfrac{k}{i+1}+\epsilon\big\}\Big).
\]
Under the Gaussian assumption~\ref{asm:gaussian} ($F(r) = \Phi((r-\mu)/\sigma)$), this gives in particular
\[
\tau_i \;\le\; \mu + \sigma\,\Phi^{-1}\!\Big(\tfrac{k}{i+1}+\epsilon\Big)
\qquad\text{with probability at least } 1 - 2e^{-2i\epsilon^2}.
\]
\end{appendixlemma}

\begin{proof}
Let $F_i$ be the empirical CDF of the first $i$ distances. The DKW inequality gives $\Pr(\sup_t |F_i(t) - F(t)| > \epsilon) \le 2e^{-2i\epsilon^2}$. On the event $\sup_t |F_i - F| \le \epsilon$ we have $F(t) - \epsilon \le F_i(t) \le F(t) + \epsilon$ for all $t$; monotonicity of $F^{-1}$ then gives $F^{-1}(u-\epsilon) \le F_i^{-1}(u) \le F^{-1}(u+\epsilon)$ for all $u \in (0,1)$. Taking $u = k/(i+1)$ and recalling $\tau_i = F_i^{-1}(k/(i+1))$ yields the two-sided bound. Under~\ref{asm:gaussian}, $F^{-1}(p) = \mu + \sigma\,\Phi^{-1}(p)$, which gives the Gaussian form.
\end{proof}

A new candidate is tested against $\tau_i$; in the Gaussian model, a representative non-near-neighbour candidate has squared distance near the mean $\mu$. This lets us define a model margin.

\begin{appendixdefinition}[Model margin $\margin_i$]\label{def:margin}
Fix $\epsilon_i \in (0,1)$ and set the sampled \kth-NN threshold upper bound
\[
K_i := \mu + \sigma\,\Phi^{-1}\!\Big(\tfrac{k}{i+1}+\epsilon_i\Big),
\qquad
\margin_i := \mu - K_i = -\sigma\,\Phi^{-1}\!\Big(\tfrac{k}{i+1}+\epsilon_i\Big).
\]
On the DKW event of Lemma~\ref{thm:threshold}, the model threshold is at most $K_i$. Thus a representative non-near-neighbour candidate at squared distance $\mu$ has model margin at least $\margin_i$. We use $\margin_i$ as the per-step representative margin in the cost model; candidate-level fluctuations are absorbed into the averaged constant $C_N$ below. The margin is positive whenever $\tfrac{k}{i+1}+\epsilon_i < \tfrac12$ (so $\Phi^{-1}(\cdot) < 0$).
\end{appendixdefinition}

\paragraph{Uniform high-probability schedule.} Fix a target failure probability $\delta \in (0,1)$ and set
\[
\epsilon_i := \sqrt{\tfrac{1}{2i}\,\ln\tfrac{2N}{\delta}}.
\]
By a union bound over $i \in \{k+1,\dots,N\}$, the event
\[
\mathcal{E}_\delta := \bigcap_{i=k+1}^{N}\Big\{\tau_i \le \mu + \sigma\,\Phi^{-1}\big(\tfrac{k}{i+1}+\epsilon_i\big)\Big\}
\]
holds with probability at least $1-\delta$. All bounds below are stated on $\mathcal{E}_\delta$.

A candidate $\xvec_j$ is pruned at level $m$ once its lower bound exceeds the threshold. A sufficient condition is that the worst-case lower-bound error falls below the margin of the candidate processed at step $i$:
\[
\|\query-\xvec_j\|^2 - \lb^m(\query,\xvec_j) \;<\; \margin_i .
\]
By Cauchy--Schwarz the left-hand side is at most four times the geometric mean of the tail energies; applying Assumptions~\ref{asm:energy} and~\ref{asm:norms},
\[
4\sqrt{R_{T(\query)}^{(m,d)}\,R_{T(\xvec_j)}^{(m,d)}}
\;\le\; 4\sqrt{\|\query\|^2\|\xvec_j\|^2\,e^{-2\alpha m/d}}
\;\le\; C_0\,e^{-\alpha m/d},\qquad C_0 := 4R^2 .
\]
Pruning is therefore guaranteed once $C_0\,e^{-\alpha m/d} < \margin_i$, i.e.
\[
e^{-\alpha m/d} < \frac{\margin_i}{C_0}
\;\iff\;
m > \frac{d}{\alpha}\,\ln\frac{C_0}{\margin_i}.
\]

\begin{appendixlemma}[Model pruning depth]\label{thm:depth}
The modelled number of dimensions $\rho_i$ processed for the candidate at step $i$ is
\[
\rho_i \;\approx\; \min\!\left\{d,\ \frac{d}{\alpha}\,\Big[\ln\frac{C_0}{\margin_i}\Big]_+\right\},
\qquad C_0 = 4R^2,\quad [x]_+ := \max\{0,x\}.
\]
\end{appendixlemma}

The total cost is dominated by the sum of the pruning dimensions over the $N$ candidates. Writing $i_0 := \min\{i \ge k+1 : \tfrac{k}{i+1}+\epsilon_i < \tfrac12\}$ for the first index at which the model margin becomes positive,
\[
\mathrm{Cost} = \sum_{i=k+1}^{N} \rho_i
\;\approx\; \sum_{i=\max\{i_0,k+1\}}^{N} \frac{d}{\alpha}\min\!\left\{\alpha,\Big[\ln\frac{C_0}{\margin_i}\Big]_+\right\}.
\]
Let $I_{C_0} := \{i \in \{\max\{i_0,k+1\},\dots,N\} : \margin_i \le C_0\}$ and $N_{C_0} := \max I_{C_0}$. Then
\[
\sum_i \Big[\ln\frac{C_0}{\margin_i}\Big]_+
= \sum_{i \in I_{C_0}} (\ln C_0 - \ln \margin_i)
= |I_{C_0}|\,\ln C_0 - \ln\!\Big(\prod_{i \in I_{C_0}} \margin_i\Big).
\]
Equivalently, define the finite-sample model constant
\begin{equation}\label{eq:CN}
C_N \;:=\; \frac{1}{N}\sum_{i=\max\{i_0,k+1\}}^{N}\min\!\left\{\alpha,\Big[\ln\frac{C_0}{\margin_i}\Big]_+\right\}.
\end{equation}
This is the average capped log-margin cost per candidate.

\begin{appendixlemma}[Model cost via averaged margins]\label{thm:margin-product}
The total computational cost is modelled by
\[
\mathrm{Cost} \;\approx\; \frac{C_N}{\alpha}\,Nd.
\]
When the cap at $d$ is inactive on the contributing window, this is equivalently
\[
C_N \;\approx\; \frac{1}{N}\Big(|I_{C_0}|\,\ln C_0 - \ln\!\prod_{i \in I_{C_0}} \margin_i\Big).
\]
\end{appendixlemma}

The exact finite-sample constant is the average in Lemma~\ref{thm:margin-product}. To see its dependence on the Gaussian model parameters, write $p_i := k/(i+1)+\epsilon_i$. When $p_i$ lies in the lower tail, the usual Gaussian-quantile approximation gives
\begin{equation}\label{eq:margin-asymp}
\margin_i = -\sigma\,\Phi^{-1}(p_i)
\;\approx\; \sigma\sqrt{2\ln(1/p_i)}.
\end{equation}
Substituting~\eqref{eq:margin-asymp} into the average shows that the constant has the representative form
\[
C_N
\;\approx\;
\frac{|I_{C_0}|}{N}\ln\frac{C_0}{\sigma}
\;-\;\zeta_N
\quad\text{(with the cap at $\alpha$ applied candidate-wise),}
\]
where
\[
\zeta_N
\;:=\;
\frac{1}{2N}\sum_{i\in I_{C_0}}\ln\!\big(2\ln(1/p_i)\big).
\]
The term $\zeta_N$ varies only through a log--log dependence on the effective tail probabilities. In finite samples we keep $C_N$ as the empirical/model constant rather than treating the endpoint approximation for $\zeta_N$ as an exact asymptotic identity.

Using Lemma~\ref{thm:margin-product}:

\begin{appendixcorollary}[Finite-sample cost form]\label{thm:final}
On $\mathcal{E}_\delta$, the modelled cost to process the candidate set is
\[
\mathrm{Cost}
\;\lesssim\; \frac{C_N}{\alpha}\,Nd,
\qquad
C_N
\;\approx\;
\frac{|I_{C_0}|}{N}\ln\frac{C_0}{\sigma}
\;-\;\zeta_N .
\]
\end{appendixcorollary}

The preceding display should be read as a finite-sample cost model. For a concrete run, the empirical counterpart is
\[
\widehat{C}_N
\;:=\;
\alpha\,\frac{\mathrm{Cost}}{Nd},
\]
the measured constant multiplying $1/\alpha$ in the dimensions-scanned fraction. The Gaussian order-statistic model predicts this constant through the average log margin $C_N$ above. Since the per-candidate work is at most $d$, the contribution of any failure of $\mathcal{E}_\delta$ is bounded by $\delta Nd$ in expectation:
\[
\mathbb{E}[\mathrm{Cost}]
\;\lesssim\;
\mathbb{E}[\mathrm{Cost}\mid\mathcal{E}_\delta]
\;+\;\delta Nd,
\]
so the same $C_N/\alpha$ scaling is unaffected by choosing $\delta$ small.

\paragraph{Comparison to the naive cost.} The naive, brute-force method computes $N$ full $d$-dimensional distances, at cost $Nd$. The bound above is smaller by a factor that scales as $\alpha$ (up to the slowly-varying and logarithmic terms), on the same high-probability event $\mathcal{E}_\delta$.

\paragraph{On the role of $\alpha$.} The parameter $\alpha$ controls the decay $e^{-\alpha m/d}$. If $\alpha \le 1$, energy decays too slowly (at $m = d/2$ the remaining energy is at least $e^{-1/2}$), leading to weak bounds and limited pruning; effective compaction corresponds to $\alpha$ comfortably greater than $1$. The high-probability analysis only replaces the expected-margin terms by their concentrated counterparts and leaves this qualitative conclusion unchanged.

\begin{appendixpropcost}[Expected refinement cost]\label{thm:cost}
Let $\phi := \mathbb{E}[\mathrm{Cost}]/(Nd)$ be the average fraction of dimensions processed per candidate. Under Assumptions~\ref{asm:energy}--\ref{asm:level},
\[
\mathbb{E}[\mathrm{Cost}] \;\lesssim\; \frac{C_N}{\alpha}\,Nd,
\qquad
\phi \;\lesssim\; \frac{C_N}{\alpha},
\]
where $C_N$ is the finite-sample constant defined in~\eqref{eq:CN}. Equivalently, \name evaluates only a fraction $C_N/\alpha$ of the $Nd$ coordinates a brute-force scan would touch---a speedup of $\alpha/C_N$ over brute force.
\end{appendixpropcost}

\begin{proof}
Combining the pruning-depth model with the definition of $C_N$,
\[
\phi = \frac{\mathbb{E}[\mathrm{Cost}]}{Nd}
\;\approx\;
\frac{1}{\alpha N}\sum_{i=\max\{i_0,k+1\}}^{N}\min\!\left\{\alpha,\Big[\ln\frac{C_0}{\margin_i}\Big]_+\right\}
\;=:\; \frac{C_N}{\alpha}.
\]
Under the Gaussian margin approximation, this is
\[
C_N \;\approx\; \frac{|I_{C_0}|}{N}\ln\frac{C_0}{\sigma} - \zeta_N,
\]
where $\zeta_N$ is the averaged log--log term above. Thus the dependence on $N$ enters through the averaged log--log term. In practice the measured dimensions-scanned fraction reflects the effective constant for the candidate stream; across the datasets in the main paper that constant is order-one and close to $1$.
\end{proof}

\subsection{Proposition \mpoodnum: Out-of-distribution cost}\label{sec:ood}

In deployment the query $\query$ and the database vectors $\{\xvec_i\}$ may compact differently under the PCA transform $T$---the query might come from a different language, sensor, or be adversarial. Let $\alpha_q$ and $\alpha_x$ denote their decay rates:
\begin{equation}\label{eq:ood-tails}
R_{T(\query)}^{(m,d)} \approx \|\query\|^2 e^{-\alpha_q m/d},
\qquad
R_{T(\xvec_i)}^{(m,d)} \approx \|\xvec_i\|^2 e^{-\alpha_x m/d}.
\end{equation}

\begin{appendixpropood}[Out-of-distribution cost]\label{thm:ood}
With asymmetric compaction rates, the effective decay rate of the lower-bound error is the \emph{arithmetic mean} $\alpha_{\mathrm{eff}} = (\alpha_q + \alpha_x)/2$, and the expected cost satisfies
\[
\mathbb{E}[\mathrm{Cost}] \;\lesssim\; \frac{C_N\,Nd}{\alpha_{\mathrm{eff}}} \;=\; \frac{2C_N\,Nd}{\alpha_q + \alpha_x},
\]
in the same model sense as \mpref{\mpcostnum}. If the OOD shift changes only the tail compaction rates, the same slowly varying margin constant $C_N$ applies; if it also changes the distance distribution, the bound holds with a modified margin constant $C_N'$ reflecting that distribution.
\end{appendixpropood}

\begin{proof}
Only the per-candidate depth changes; the threshold, margin, and aggregation arguments are unchanged. The Cauchy--Schwarz remainder obeys, after substituting the asymmetric tails~\eqref{eq:ood-tails} and bounding norms by Assumption~\ref{asm:norms},
\[
\begin{aligned}
&4\sqrt{R_{T(\query)}^{(m,d)}\,R_{T(\xvec_j)}^{(m,d)}}\\
&\quad \le 4R^2\sqrt{e^{-\alpha_q m/d}\,e^{-\alpha_x m/d}}\\
&\quad = 4R^2\,e^{-(\alpha_q + \alpha_x)\,m/(2d)}\\
&\quad = C_0\,e^{-\alpha_{\mathrm{eff}}\,m/d},
\end{aligned}
\]
with $\alpha_{\mathrm{eff}} = (\alpha_q + \alpha_x)/2$. Solving $C_0\,e^{-\alpha_{\mathrm{eff}} m/d} < \margin_i$ gives the capped per-candidate depth $\rho_i \approx \min\{d,(d/\alpha_{\mathrm{eff}})[\ln(C_0/\margin_i)]_+\}$. The threshold, margin, and aggregation arguments above are unchanged with $\alpha$ replaced by $\alpha_{\mathrm{eff}}$, yielding the stated cost.
\end{proof}

\begin{appendixremark}[Graceful degradation]
Even a fully out-of-distribution query with no compaction ($\alpha_q \to 0$) leaves $\mathbb{E}[\mathrm{Cost}] \lesssim 2C_N\,Nd/\alpha_x$: \name still benefits from the database compaction alone, a speedup of roughly $\alpha_x/(2C_N)$ over brute force when the margin constant is unchanged. Performance degrades gracefully rather than collapsing, as long as the database vectors are well-compacted.
\end{appendixremark}

\subsection{Proposition \mpepsnum: The $\varepsilon$-relaxation}\label{sec:eps}

Setting $\varepsilon = 1$ uses the exact Cauchy--Schwarz correction $2C^m$ and preserves recall. That worst-case correction is attained only at the adversarial residual alignment $\rho_m = -1$; on real, isotropised residuals the typical alignment is $\rho_m \approx 0$, where the correction is already a factor of two smaller. The $\varepsilon$-relaxation exploits this gap: it discounts the correction by a factor $\varepsilon \in (0,1]$, pruning earlier at a controlled, quantifiable cost in recall. We first record the exact gap identity, then state the isotropy assumption that makes the typical case precise, and finally turn it into a cost bound that mirrors \mpref{\mpcostnum}.

The $\varepsilon$-relaxed bound is
\[
\lb^{m}_{\varepsilon}(\query,\xvec) \;=\; A^{m} - 2\varepsilon\,C^{m}, \qquad C^{m} := \sqrt{R_{T(\query)}^{(m,d)}\,R_{T(\xvec)}^{(m,d)}},
\]
where $A^m = R_{T(\query)}^{(0,d)} + R_{T(\xvec)}^{(0,d)} - 2 p^{(0,m)}(\query,\xvec)$. Define the residual inner product
\[
\pi_m \;=\; \sum_{j > m} T(\query)_j\,T(\xvec)_j
\]
and its normalised value
\[
\rho_m \;:=\; \frac{\pi_m}{C^m} \;\in\; [-1, 1],
\]
where the bounds follow from Cauchy--Schwarz applied to the residual.

\begin{appendixlemma}[Gap identity]\label{lem:gap-identity}
For every level $m$,
\[
\|\query - \xvec\|^2 - \lb^{m}_{\varepsilon}(\query,\xvec) \;=\; 2 C^{m}(\varepsilon - \rho_m).
\]
\end{appendixlemma}

\begin{proof}
Direct expansion:
\begin{align*}
\|\query-\xvec\|^2 &= R_{T(\query)}^{(0,d)} + R_{T(\xvec)}^{(0,d)} - 2\sum_{j=1}^{d} T(\query)_j T(\xvec)_j \\
&= A^m - 2\pi_m.
\end{align*}
Subtracting the relaxed bound,
\[
\begin{aligned}
\|\query-\xvec\|^2 - \lb^{m}_{\varepsilon}(\query,\xvec)
&= (A^m - 2\pi_m) - (A^m - 2\varepsilon C^m)\\
&= 2(\varepsilon C^m - \pi_m)\\
&= 2 C^m(\varepsilon - \rho_m). \qedhere
\end{aligned}
\]
\end{proof}

At $\varepsilon=1$ and worst case $\rho_m = -1$ the gap is $4 C^m$, matching the strict cost bound; at $\varepsilon=1$ and typical $\rho_m \approx 0$ the gap is $2 C^m$, a factor of $2$ tighter.

\begin{appendixassumption}[Residual isotropy]\label{asm:isotropy}
For non-near-neighbour pairs $(\query, \xvec)$, the residual coordinates $(T(\query)_j, T(\xvec)_j)_{j > m}$ are independent and uniformly distributed on the sphere in $\mathbb{R}^{d-m}$ (independently for $\query$ and $\xvec$). Consequently, by standard high-dimensional concentration,
\[
\rho_m \;\sim\; \mathcal{N}\!\left(0,\,\tfrac{1}{d-m}\right) \quad \text{approximately.}
\]
\end{appendixassumption}

\begin{appendixremark}
Assumption~\ref{asm:isotropy} is realised by the intra-level isotropization step in the main paper's variance-shaping transform. It applies only to non-near-neighbour pairs; for true $k$-NN pairs the residuals are correlated (the nearness manifests in the residual), so $\rho_m$ for true neighbours concentrates strictly above $0$. The recall loss at $\varepsilon < 1$ arises precisely from neighbour pairs with $\rho_m > \varepsilon$.
\end{appendixremark}

\begin{appendixlemma}[Pruning depth, $\varepsilon$-relaxed]\label{lem:depth-eps}
Under Assumptions~\ref{asm:energy}--\ref{asm:isotropy}, for non-near-neighbour candidates and a margin $\margin > 0$, the pruning depth at the $\varepsilon$-relaxed bound is
\[
d^{(\varepsilon)}(\margin) \;\approx\; \frac{d}{\alpha}\,\ln\!\frac{2 R^2}{\margin} \;+\; \frac{d}{\alpha}\,\ln \varepsilon.
\]
\end{appendixlemma}

\begin{proof}
Under~\ref{asm:isotropy}, $\rho_m$ concentrates at $0$ with deviation $1/\sqrt{d-m}$. Substituting $\rho_m \approx 0$ in Lemma~\ref{lem:gap-identity},
\[
\|\query - \xvec\|^2 - \lb^{m}_{\varepsilon}(\query, \xvec) \;\approx\; 2\varepsilon\,C^m.
\]
Applying Assumption~\ref{asm:energy} to both factors of $C^m$ and Assumption~\ref{asm:norms},
\[
2\varepsilon\,C^m \;\approx\; 2\varepsilon\,\|\query\|\,\|\xvec\|\,e^{-\alpha m/d} \;\le\; 2\varepsilon\,R^2\,e^{-\alpha m/d}.
\]
Pruning occurs at the smallest $m$ for which this is at most $\margin$:
\[
2\varepsilon R^2 e^{-\alpha m/d} \;\le\; \margin \;\iff\; m \;\ge\; \frac{d}{\alpha}\ln\!\frac{2\varepsilon R^2}{\margin} \;=\; \frac{d}{\alpha}\ln\!\frac{2 R^2}{\margin} \;+\; \frac{d}{\alpha}\ln\varepsilon.
\qedhere
\]
\end{proof}

The first term is the typical-case pruning depth at $\varepsilon = 1$, with prefactor $2R^2$ rather than the strict-mode prefactor $C_0 = 4R^2$. The two prefactors differ by a constant inside the log and are absorbed into the same slowly-varying constant as in \mpref{\mpcostnum}.

\begin{appendixpropeps}[Cost under $\varepsilon$-relaxation]\label{thm:eps}
Under Assumptions~\ref{asm:energy}--\ref{asm:isotropy} and $\varepsilon \in (1/e,\,1]$, the expected refinement cost of \name with the $\varepsilon$-relaxed bound satisfies
\[
\mathbb{E}[\mathrm{Cost}_\varepsilon] \;\lesssim\; \frac{Nd\,(c_{\mathrm{typ}}+\ln \varepsilon)}{\alpha},
\]
where $c_{\mathrm{typ}}$ is the typical-residual version of the constant $C_N$ (with prefactor $2R^2$ instead of the strict worst-case $4R^2$). Since the empirical typical-case constant is close to $1$, the simplified model is $\mathbb{E}[\mathrm{Cost}_\varepsilon] \lesssim Nd(1+\ln\varepsilon)/\alpha$, equivalently $\alpha_{\mathrm{eff}}(\varepsilon) \approx \alpha/(1+\ln\varepsilon)$. For $\varepsilon < 1$, each candidate is pruned $(d/\alpha)\ln(1/\varepsilon)$ dimensions earlier on average than at $\varepsilon = 1$ until the cap at zero dimensions becomes active.
\end{appendixpropeps}

\begin{proof}
The argument mirrors the cost proof, with the pruning depth of Lemma~\ref{thm:depth} replaced by Lemma~\ref{lem:depth-eps}. We work on the same high-probability event $\mathcal{E}_\delta$ (Lemma~\ref{thm:threshold}); the threshold and margin analysis are unchanged. Substituting the margin asymptotic~\eqref{eq:margin-asymp} into Lemma~\ref{lem:depth-eps} and aggregating as in \mpref{\mpcostnum}, the additive $(d/\alpha)\ln\varepsilon$ term carries through the sum to give
\[
\mathbb{E}[\mathrm{Cost}_\varepsilon] \;\lesssim\; \frac{Nd}{\alpha}\,(c_{\mathrm{typ}} + \ln\varepsilon),
\]
where $c_{\mathrm{typ}} = \ln(2R^2/\sigma) - \zeta_N$ is the typical-residual counterpart of $C_N$, shifted only by $\ln 2$ (the typical-case prefactor is $2R^2$ rather than $C_0 = 4R^2$) and inheriting the same log--log behaviour. Since $c_{\mathrm{typ}}$ is observed to be $\approx 1$,
\[
\mathbb{E}[\mathrm{Cost}_\varepsilon] \;\lesssim\; \frac{Nd\,(1 + \ln \varepsilon)}{\alpha}. \qedhere
\]
\end{proof}

\begin{appendixremark}[Validity range]
The asymptotic requires $c_{\mathrm{typ}} + \ln \varepsilon > 0$. Setting $c_{\mathrm{typ}} = 1$ gives the conventional threshold $\varepsilon > 1/e \approx 0.37$; the actual threshold is dataset-dependent. Below it, the formula predicts a negative pruning depth---an artifact: the bound clears the margin already at $m = 0$, and the cost is governed by other factors (heap warm-up, fixed per-candidate overhead). In this regime $\varepsilon$ should be tuned empirically.
\end{appendixremark}

\begin{appendixremark}[Additive vs.\ multiplicative]
The $\varepsilon$-relaxation contributes \emph{additively} to the per-candidate pruning depth, in contrast to the multiplicative $1/\alpha$ scaling of the base model. A moderate $\varepsilon = 0.7$ shifts the prefactor by $\ln 0.7 \approx -0.36$, enough to produce a noticeable QPS gain without changing the leading $1/\alpha$ behaviour, while preserving near-perfect recall.
\end{appendixremark}

\mpref{\mpepsnum} characterises the speed side of the $\varepsilon$ knob. Pruning at $\varepsilon < 1$ discards true $k$-nearest-neighbour pairs whose residual ratio $\rho_m$ exceeds $\varepsilon$. The recall loss thus depends on the dataset-specific distribution of $\rho_m$ for true $k$-NN pairs, which is measurable per dataset (see the $\varepsilon$ ablation in the main paper). We do not attempt a closed-form recall bound; the $\rho_m$ distribution for $k$-NN pairs concentrates strictly above $0$ (these pairs are correlated, breaking the isotropy of A5), and recall begins to degrade once $\varepsilon$ falls below the bulk of this distribution.

\end{document}